\documentclass{article}

\usepackage{booktabs} 

\usepackage{hyperref}

\usepackage[accepted]{icml2024}

\usepackage{amsmath}
\usepackage{amssymb}
\usepackage{mathtools}
\usepackage{amsthm}

\usepackage[capitalize,noabbrev]{cleveref}

\usepackage{url}
\usepackage{xcolor}         
\usepackage{colortbl}

\usepackage{graphicx}
\usepackage{subcaption}
\usepackage{graphicx}
\usepackage{wrapfig}
\usepackage[inline]{enumitem}

\theoremstyle{plain}

\theoremstyle{definition}

\theoremstyle{remark}

\usepackage[textsize=tiny]{todonotes}

\icmltitlerunning{Lookbehind-SAM: $k$ steps back, 1 step forward}

\begin{document}

\twocolumn[
\icmltitle{Lookbehind-SAM: $k$ steps back, 1 step forward}




\begin{icmlauthorlist}

\icmlauthor{Gonçalo Mordido}{mila,poly}
\icmlauthor{Pranshu Malviya}{mila,poly}
\icmlauthor{Aristide Baratin}{samsung}
\icmlauthor{Sarath Chandar}{mila,poly,ccai}
\end{icmlauthorlist}

\icmlaffiliation{poly}{Polytechnique Montreal}
\icmlaffiliation{mila}{Mila - Quebec AI Institute}
\icmlaffiliation{samsung}{Samsung SAIT AI Lab Montreal}
\icmlaffiliation{ccai}{Canada CIFAR AI Chair}

\icmlcorrespondingauthor{Gonçalo Mordido}{goncalomordido@gmail.com}

\icmlkeywords{Machine Learning, ICML}

\vskip 0.3in
]



\printAffiliationsAndNotice{}  

\begin{abstract}
Sharpness-aware minimization (SAM) methods have gained increasing popularity by formulating the problem of minimizing both loss value and loss sharpness as a minimax objective. 
In this work, we increase the efficiency of the maximization and minimization parts of SAM's objective to achieve a better loss-sharpness trade-off.
By taking inspiration from the Lookahead optimizer, which uses multiple descent steps ahead, we propose Lookbehind, which performs multiple ascent steps behind to enhance the maximization step of SAM and find a worst-case perturbation with higher loss. Then, to mitigate the variance in the descent step arising from the gathered gradients across the multiple ascent steps, we employ linear interpolation to refine the minimization step. 
Lookbehind leads to a myriad of benefits across a variety of tasks. Particularly, we show increased generalization performance, greater robustness against noisy weights, as well as improved learning and less catastrophic forgetting in lifelong learning settings. Our code is available at \url{https://github.com/chandar-lab/Lookbehind-SAM}.
\end{abstract}

\section{Introduction}

Improving the optimization methods used in deep learning is a crucial step to enhance the performance of current models.
Notably, building upon the long-recognized connection between the flatness of the loss landscape and generalization \citep{hochreiter1994simplifying,keskar2016large,DR17,neyshabur2017exploring,DBLP:conf/uai/IzmailovPGVW18}, sharpness-aware training methods have gained recent popularity due to their ability to significantly improve generalization performance compared to minimizing the empirical risk using stochastic gradient descent (SGD). Particularly, sharpness-aware minimization (SAM) \citep{foret2021sharpnessaware} was recently proposed as an effective means to simultaneously minimize both loss value and loss sharpness during training. Given a neural network with parameters $\phi$, some loss function $L(\phi)$, SAM seeks parameters in flat regions using a minimax optimization: 
\begin{equation} 
\min_{\phi} \max_{\|\epsilon\|_2 \leq \rho} L(\phi + \epsilon) \, ,
\label{eq:sam}
\end{equation} 
where worst-case perturbations $\epsilon$ are applied to parameters $\phi$, with the distance between original and perturbed parameters being controlled by $\rho$. SAM approximates the maximization step by first performing a single gradient ascent step and then using the gradient of the loss to do a single descent step from the original solution. 
This leads to finding a low-loss parameter configuration $\phi$ such that the loss is also low in the neighborhood $\rho$ which will lead to flatter solutions.
Several follow-up methods have emerged to further enhance its performance \citep{kwon2021asam, zhuang2022surrogate, kim2022fisher}  and reduce its computation overhead \citep{du2022sharpnessaware, du2022efficient, liu2022towards}.

\begin{figure}
  \begin{center}
    \includegraphics[width=0.4\textwidth]{{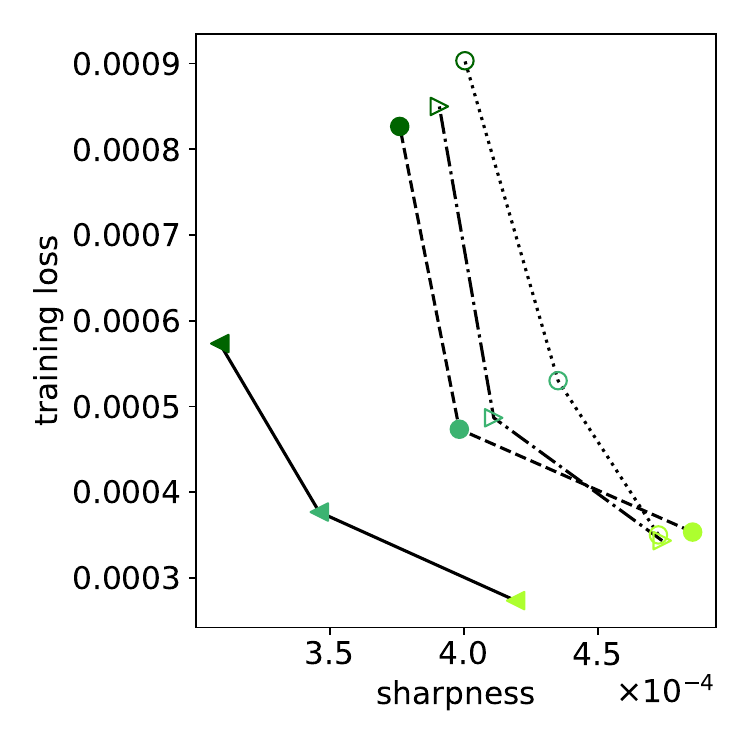}}
  \end{center}
  \begin{center}
    \includegraphics[width=0.41\textwidth]{{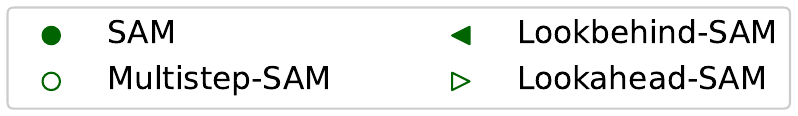}}
  \end{center}
  \caption{Loss and sharpness trade-off using ResNet-34 trained on CIFAR-10. Darker shades indicate training with higher neighborhood sizes $\rho \in \{0.05, 0.1, 0.2\}$.}
  \label{fig:loss_sharpness_trade_off}
  \vspace{-10.5px}
\end{figure}

Despite the recent success, improving upon SAM requires a delicate balance between loss value and sharpness. 
Ideally, the optimization process would converge towards minima that offer a favorable compromise between these two aspects,  thereby leading to high generalization performance. However, naively increasing the neighborhood size $\rho$ used to find the perturbed solutions in SAM leads to a considerable increase in training loss, despite improving sharpness (Figure \ref{fig:loss_sharpness_trade_off}, full circles). In other words, putting too much emphasis on finding the worst-case perturbation is expected to bias convergence to flat but high-loss regions and negatively impact generalization performance. 

Instead of performing a single ascent step akin to SAM, performing multiple ascent steps is a promising way of increasing the neighborhood region to find perturbed solutions, further reducing sharpness. However, this is not what is observed empirically (Figure \ref{fig:loss_sharpness_trade_off}, empty circles). In fact, previous works \citep{foret2021sharpnessaware,andriushchenko2022towards} have shown that such a multistep variant may hurt performance. 
A possible cause is the increased gradient instability originating from moving farther away from our original solution \citep{liu2022random}. 
Note that such instability may also be present when using a high $\rho$, even in single-ascent step SAM. In this case,  applying a variance reduction technique such as Lookahead \citep{zhang2019lookahead} with SAM as inner optimizer may help mitigate the performance loss when using larger $\rho$. However, as we demonstrate in our experiments, this is also not helpful (Figure \ref{fig:loss_sharpness_trade_off}, empty triangles).

In this work, we present a novel optimization method, Lookbehind, that leverages the benefits of multiple ascent steps and variance reduction to improve the efficiency of the maximization and
minimization parts of \eqref{eq:sam}.
By successfully reducing both loss and sharpness across small and large neighborhood sizes (Figure \ref{fig:loss_sharpness_trade_off}, full triangles), Lookbehind achieves the best loss-sharpness trade-off.

In practice, improving the loss and sharpness trade-off results in a myriad of benefits across several training regimes. Particularly, when applying Lookbehind to SAM and ASAM, we show an improvement in terms of generalization performance across several models and datasets. 
Moreover, models trained with Lookbehind have increased robustness against noisy weights at inference time. 
Lastly, we evaluate Lookbehind in the context of lifelong learning and show an improvement both in terms of learning and catastrophic forgetting on multiple models and datasets.

\section{Sharpness-aware minimization}

Our method, Lookbehind, builds upon sharpness-aware minimization (SAM) methods with the goal of solving the inner maximization problem of SAM more accurately while stabilizing the outer minimization part of SAM's objective. We will start by briefly introducing the sharpness-aware minimization methods used throughout the paper.

To solve the problem in \eqref{eq:sam} using standard stochastic gradient methods, SAM \citep{foret2021sharpnessaware} proposes to estimate the gradient of the minimax objective in two steps. The first step is to approximate the inner maximization $\epsilon(\phi)$ using one step of gradient ascent; the second is to compute the loss gradient at the perturbed parameter $\phi+\epsilon(\phi)$. This leads to the following parameter update:
\begin{align}
\phi_t = \phi_{t-1} - \eta \nabla_{\phi} L(\phi_{t-1}+\epsilon(\phi_{t-1})) \, , \\  \epsilon(\phi) := \rho\frac{\nabla L(\phi)}{||\nabla L(\phi)||_2} \, .
\end{align}

Several follow-up sharpness-aware methods have been proposed to further improve upon the original formulation. Notably, a conceptual drawback of SAM is the use of a fixed-radius Euclidean ball as maximization neighborhood, which is sensitive to re-parametrizations such as weight re-scaling \citep{dinh2017sharp,Stutz_2021_ICCV}. To address this problem, ASAM \citep{kwon2021asam} was proposed as an adaptive version of SAM, which redefines the maximization neighborhood in \eqref{eq:sam}  as component-wise normalized balls $\|\epsilon / |\phi|\|_2 \leq \rho$. This leads to the following component-wise rescaling:
\begin{align}
\epsilon(\phi) := \rho \: \frac{T_\phi^2(\nabla L(\phi))}{||T_\phi (\nabla L(\phi))||_2} \, ,
\label{eq:ASAM}
\end{align}
where $T_\phi(v) := \phi \odot v $ denotes the component-wise multiplication operator associated to $\phi$. In what follows, we use both SAM and ASAM as our baseline sharpness-based learning methods.

\section{Lookbehind optimizer}
\label{sec:LB+SAM}

\begin{figure*}[h]
     \centering
     \begin{subfigure}[b]{0.44\textwidth}
         \centering
         \includegraphics[width=\textwidth]{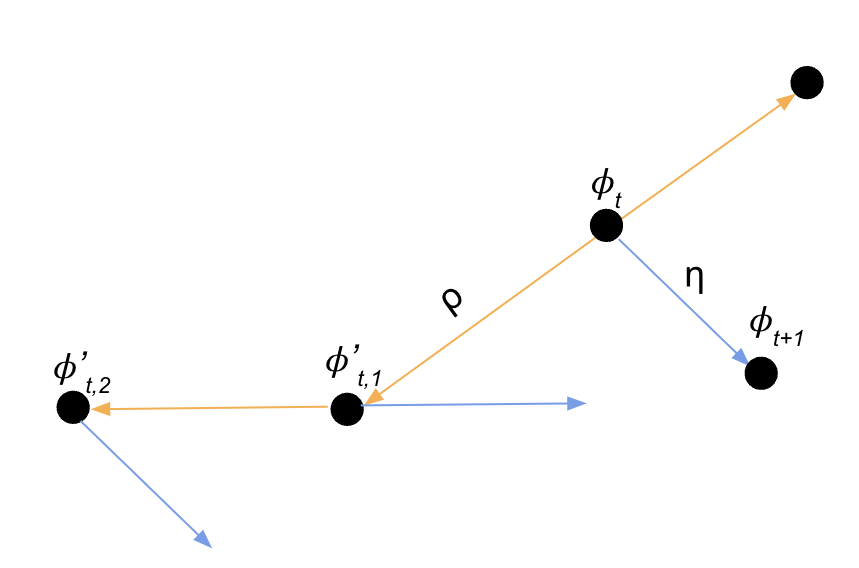}
         \caption{Multistep-SAM.}
     \end{subfigure}
     \hfill
     \begin{subfigure}[b]{0.54\textwidth}
         \centering
         \includegraphics[width=\textwidth]{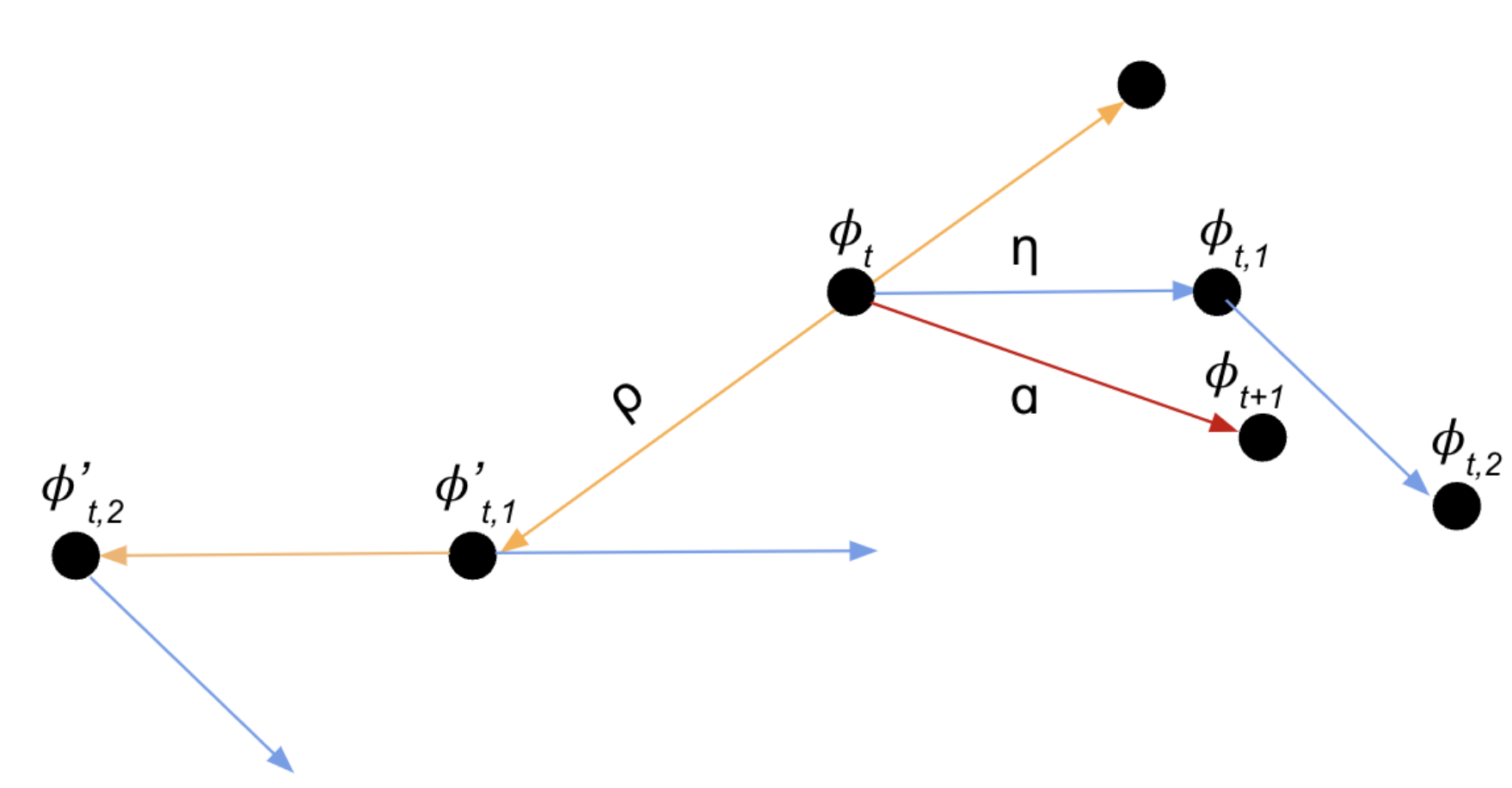}
         \caption{Lookbehind-SAM (ours).}
     \end{subfigure}
        \caption{Illustration of Multistep-SAM (a) and Lookbehind-SAM (b) using $k=2$.}
        \label{fig:sams}
\end{figure*}

Our algorithm, Lookbehind (-SAM), presents a novel way to improve the solution found by SAM's objective \eqref{eq:sam}. 
The intuition of Lookbehind is two-fold. First, we improve the maximization part of SAM’s objective by performing multiple ascent steps to find a worst-case weight perturbation that has a higher loss than the original, single-step SAM within a given neighborhood of the original point. We refer to such maximization of the loss as we perform multiple ascent steps in SAM as looking behind. In other words, we are looking behind in the sense that we are climbing the loss landscape. (This term is inspired by the Lookahead optimizer \citep{zhang2019lookahead}, where looking ahead refers to the minimization of the loss as they perform multiple descent steps.)

Second, to improve the minimization part of SAM’s objective, we reduce the variance derived from the multiple ascent steps by aggregating the gradients along the way for the descent step and performing linear interpolation in the parameter space. 
This results in an alleviation of the instability that arises from (i) performing multiple ascent steps due to the various gradients gathered in the ascent phase not being aligned with each other and (ii) the substantial departure away from the original point as performing ascent steps, which negatively impacts SAM's minimization objective and consequent loss-sharpness trade-off (Figure \ref{fig:loss_sharpness_trade_off}). Lookbehind combines instead the gradients computed at intermediate distances, improving upon the multiple ascent step variant of SAM (Multistep-SAM). A visual comparison between Multistep-SAM and Lookbehind is illustrated in Figure \ref{fig:sams}.

While Multistep-SAM performs $k$ ascent steps ($\phi'_{t, 1}$, $\cdots$, $\phi'_{t, k}$) and uses the gradient from the last step ($\phi'_{t,k}$) for the final update, 
Lookbehind uses slow weights ($\phi_{t}$, $\phi_{t+1}, \cdots$) and fast weights 
\begin{figure}
\vspace{-12px}
\begin{minipage}{0.51\textwidth}
\begin{algorithm}[H]
\caption{Lookbehind-SAM}
\begin{algorithmic}[1]
\REQUIRE Parameters $\phi_0$, loss $L$, inner steps $k$, slow and fast weights step
sizes $\alpha$ and $\eta$, neighborhood size $\rho$, training set $D$
\FOR{$t = 1, 2, \ldots$}
\STATE $\phi_{t,0} \gets \phi_{t-1}$
\STATE $\phi_{t,0}' \gets \phi_{t-1}$
\STATE Sample mini-batch $d \sim D$
\FOR{$i = 1, 2, \ldots, k$}
\STATE $\epsilon \gets \rho \dfrac{\nabla L_d(\phi_{t,i-1}')}{\|\nabla L_d(\phi_{t,i-1}')\|_2}$
\STATE $\phi_{t,i}' \gets \phi_{t,i-1}' + \epsilon$
\STATE $\phi_{t,i} \gets \phi_{t,i-1} - \eta \nabla_{L_{d}} (\phi_{t,i}')$
\ENDFOR
\STATE $\phi_{t} \gets \phi_{t-1} + \alpha(\phi_{t,k} - \phi_{t-1})$
\ENDFOR
\STATE {\bfseries return} $\phi$
\end{algorithmic}
\label{alg:lookbehind}
\end{algorithm}
\end{minipage}
\end{figure}
($\phi_{t, 1}$, $\cdots$, $\phi_{t, k}$), where fast weights are updated using the gradients from $k$ ascent SAM steps.
Then, the slow weights are updated toward the fast weights through linear interpolation. Even though both methods entail the same number of gradient computations, Lookbehind has a stabilizing effect over Multistep-SAM by combining the gradient information.

The pseudo-code for Lookbehind is in Algorithm \ref{alg:lookbehind}. After synchronizing the fast weights (line 2) and the perturbed weights (line 3), we sample a minibatch (line 4) and perform $k$ ascent steps of SAM by preserving the previously perturbed slow weights (line 7) and introducing further perturbations in the subsequent inner step (line 6); corresponding descent steps are tracked and the fast weights are updated accordingly (line 8). After $k$ steps, a linear interpolation of the fast and slow weights is conducted (line 10). We note that the slow weight step size, $\alpha$, can be set in an adaptive manner during training, without requiring hyperparameter tuning (see \cref{sec:adaptive_alpha}).

\begin{table*}[t]
\caption{Generalization performance (validation acc. \%) of the different methods on several models and datasets.
}
\begin{center}
\resizebox{0.8\textwidth}{!}{
\begin{tabular}{|l|c|c||c|c||c|}
\hline
Dataset & \multicolumn{2}{c||}{CIFAR-10} & \multicolumn{2}{c||}{CIFAR-100} & ImageNet\\
Model & ResNet-34 & WRN-28-2 & ResNet-50 & WRN-28-10 & ResNet-18\\
\hline\hline
SGD & $95.84_{\pm .13}$ & $93.58_{\pm .11}$ & $74.35_{\pm 1.23}$ & $78.80_{\pm .08}$ & $69.91_{\pm .04}$\\
Lookahead-SGD & $95.59_{\pm .21}$ & $94.01_{\pm .02}$  & $75.96_{\pm .12}$ & $78.53_{\pm .18}$ & $69.63_{\pm .12}$\\
\hline
SAM & $95.80_{\pm .07}$ & $93.97_{\pm .20}$ & $76.57_{\pm .59}$ & $80.50_{\pm .06}$ & $70.01_{\pm .06}$\\
Multistep-SAM & $95.72_{\pm .15}$ & $94.39_{\pm .09}$ & $77.03_{\pm .65}$ & $80.55_{\pm .06}$ & $69.92_{\pm .07}$\\
\quad + average grads & $95.74_{\pm .25}$ & $94.55_{\pm .22}$ & $76.97_{\pm .57}$ & $80.58_{\pm .21}$ & $70.01_{\pm .07}$\\
Lookahead-SAM & $95.80_{\pm .11}$ & $93.97_{\pm .17}$ & $76.16_{\pm .98}$ & $80.09_{\pm .10}$ & $69.99_{\pm .07}$\\
\textbf{Lookbehind-SAM} & $\pmb{96.27_{\pm .07}}$  & $\pmb{94.81_{\pm .22}}$  & $\pmb{78.62_{\pm .48}}$ & $\pmb{80.99_{\pm .02}}$ & $\pmb{70.16_{\pm .08}}$\\
\hline
ASAM & $96.32_{\pm .02}$ & $94.41_{\pm .09}$  & $78.62_{\pm .67}$ & $81.67_{\pm .28}$ & $70.15_{\pm .06}$\\
Multistep-ASAM & $95.91_{\pm .14}$ & $95.06_{\pm .15}$ & $77.81_{\pm .52}$ & $81.67_{\pm .06}$ & $70.06_{\pm .01}$\\
\quad + average grads & $95.91_{\pm .24}$ & $94.92_{\pm .09}$ & $78.39_{\pm .52}$ & $81.35_{\pm .36}$ & $70.11_{\pm .03}$\\
Lookahead-ASAM & $96.01_{\pm .15}$  & $94.28_{\pm .04}$  & $77.55_{\pm 1.10}$  & $80.97_{\pm .17}$ & $70.00_{\pm .11}$\\
\textbf{Lookbehind-ASAM} & $\pmb{96.54_{\pm .21}}$  & $\pmb{95.23_{\pm .01}}$  & $\pmb{78.86_{\pm .29}}$ & $\pmb{82.16_{\pm .09}}$ & $\pmb{70.23_{\pm .22}}$\\
\hline
\end{tabular}}
\end{center}
\label{tab:generalization}
\end{table*}

\section{Experimental results}

In this section, we start by introducing our baselines (Section \ref{sec:baselines}), and then we conduct several experiments to showcase the benefits of achieving a better sharpness-loss trade-off in SAM methods.
Particularly, we test the generalization performance on several models and datasets (Section \ref{sec:generalization_performance}) and analyze the loss landscapes at the end of training in terms of sharpness (Section \ref{sec:sharpness}). Then, we study the robustness provided by the different methods in noisy weight settings (Section \ref{sec:robustness}). Lastly, we assess continual learning in sequential training settings (Section \ref{sec:lifelong_learning}).

For the following experiments, we use residual networks (ResNets) \citep{he2016deep} and wide residual networks (WRN) \citep{zagoruyko2016wide} models trained from scratch on CIFAR-10, CIFAR-100 \citep{krizhevsky2009learning}, and ImageNet \citep{deng2009imagenet}. 
 We report the mean and standard deviation over 3 different seeds throughout the paper unless noted otherwise. Additional training and hyperparameter search details are provided in Appendices \ref{sec:app_training_details} and \ref{sec:hyperparameter_tuning}.

\subsection{Baselines}
\label{sec:baselines}

On top of the previously discussed Lookbehind-SAM, our algorithm can be easily combined with ASAM by using the component-wise rescaling \eqref{eq:ASAM} in the inner loop updates. We call this variant Lookbehind-ASAM. Additionally to SGD and vanilla SAM/ASAM, we compare Lookbehind-SAM/ASAM to the following methods:
\begin{enumerate*}[label=(\roman*)]
    \item \emph{Multistep-SAM/ASAM}, which performs multiple ascent steps to SAM/ASAM with the final update using the gradient from the last step,
    \item \emph{Multistep-SAM/ASAM with gradient averaging}, which applies the average of the accumulated gradients for the final update \citep{kim2023exploring},
    \item \emph{Lookahead-SAM/ASAM}, which uses Lookahead with sharpness-aware methods by applying single-step SAM/ASAM as the inner optimizer (more details are provided in Appendix \ref{sec:lookahead_sam}), and
    \item \emph{Lookahead-SGD}, which applies the Lookahead optimizer to SGD, as originally proposed by \citet{zhang2019lookahead}. 
\end{enumerate*}

\subsection{Generalization performance}
\label{sec:generalization_performance}

We start by reporting the generalization performance on several models and datasets in Table \ref{tab:generalization}. We observe that models trained with Lookbehind achieve the best generalization performance across all architectures and datasets. This is observed for both SAM and ASAM. Moreover, we see the Lookbehind-SAM/ASAM variants always outperform Lookahead-SGD, which further validates applying Lookbehind to sharpness-aware minimization methods. Importantly, we note that Lookbehind is the only method to outperform vanilla SAM and ASAM on ImageNet. The improvement of the loss-sharpness trade-off by Lookbehind leads to a myriad of additional benefits, as shown next.

\subsection{Sharpness across large neighborhood regions}
\label{sec:sharpness}

We move on to analyzing the sharpness of the minima found at the end of training for each method. To do this, we measure the sharpness of the trained models using SAM's $m$-sharpness \citep{foret2021sharpnessaware} by computing 
\begin{equation}
    \dfrac{1}{n} \sum_{M \in D} \max_{\|\epsilon\|_2 \leq r} \dfrac{1}{m} \sum_{s \in M} L_s(\phi + \epsilon) - L_s(\phi) \, ,
    \label{eq:m_sharpness_sam}
 \end{equation}
where $D$ represents the training dataset, which is composed of $n$ minibatches $M$ of size $m$. Note that $m$-sharpness can also derived from ASAM's objective by computing
\begin{equation}
    \dfrac{1}{n} \sum_{M \in D} \max_{\|\epsilon / |\phi| \|_2 \leq r} \dfrac{1}{m} \sum_{s \in M} L_s(\phi + \epsilon) - L_s(\phi) \, .
    \label{eq:m_sharpness_asam}
\end{equation}
To avoid ambiguity, we denote the radius used by $m$-sharpness as $r$. Instead of only measuring sharpness in close vicinity to the found solutions, \textit{i.e.} using $r=0.05$ as in Figure \ref{fig:loss_sharpness_trade_off}, we vary the radius $r$ over which $m$-sharpness is calculated. Particularly, we iterate over $r \in \{0.05, 0.5, 1.0, \ldots, 5.0\}$ for SAM and $r \in \{0.5, 1.0, \ldots, 5.0\}$ for ASAM. 


\begin{figure*}[t]
     \centering
     \begin{subfigure}[b]{0.32\textwidth}
         \centering
         \includegraphics[width=\textwidth]{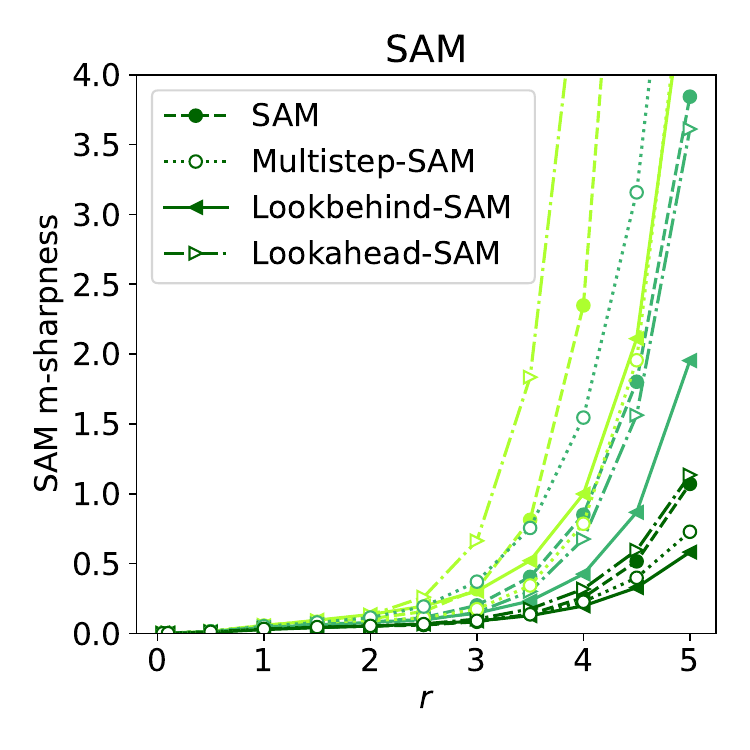}
     \end{subfigure}
     \begin{subfigure}[b]{0.32\textwidth}
         \centering
         \includegraphics[width=\textwidth]{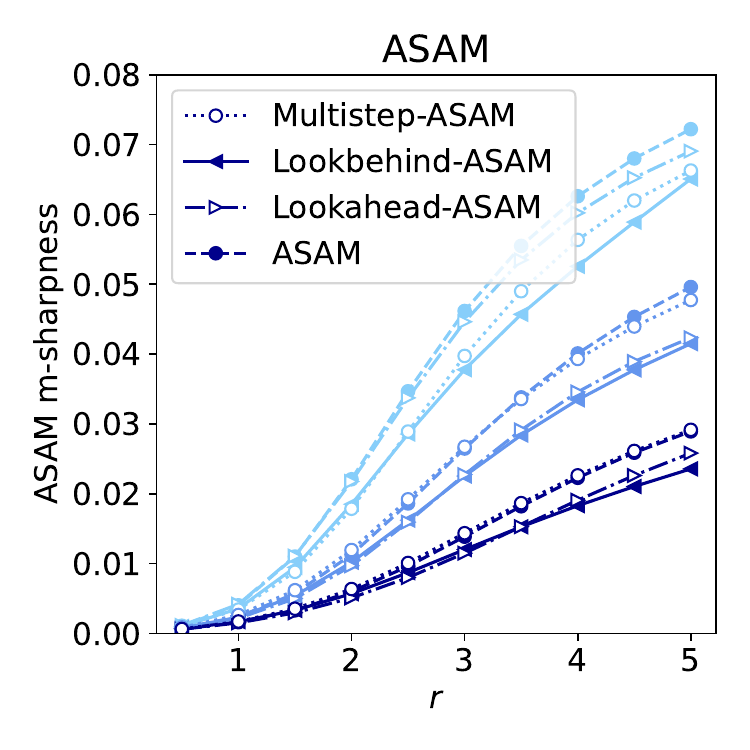}
     \end{subfigure}
     \hfill
        \caption{$m$-sharpness over multiple radius $r$ using ResNet-34 trained on CIFAR-10. Darker shades indicate training with higher neighborhood sizes $\rho \in \{0.05, 0.1, 0.2\}$ for SAM and $\rho \in \{0.5, 1.0, 2.0\}$ for ASAM. Lower sharpness is better.
        }
        \label{fig:sharpness_rho}
\end{figure*}

\begin{figure*}[t]
     \centering
     \begin{subfigure}[b]{0.32\textwidth}
         \centering
         \includegraphics[width=\textwidth]{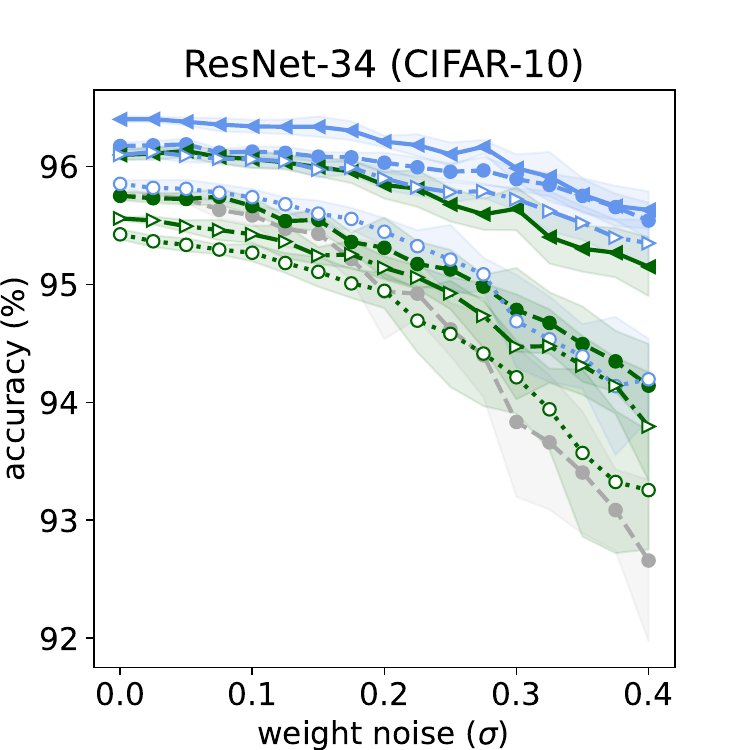}
     \end{subfigure}
     \hfill
     \begin{subfigure}[b]{0.32\textwidth}
         \centering
         \includegraphics[width=\textwidth]{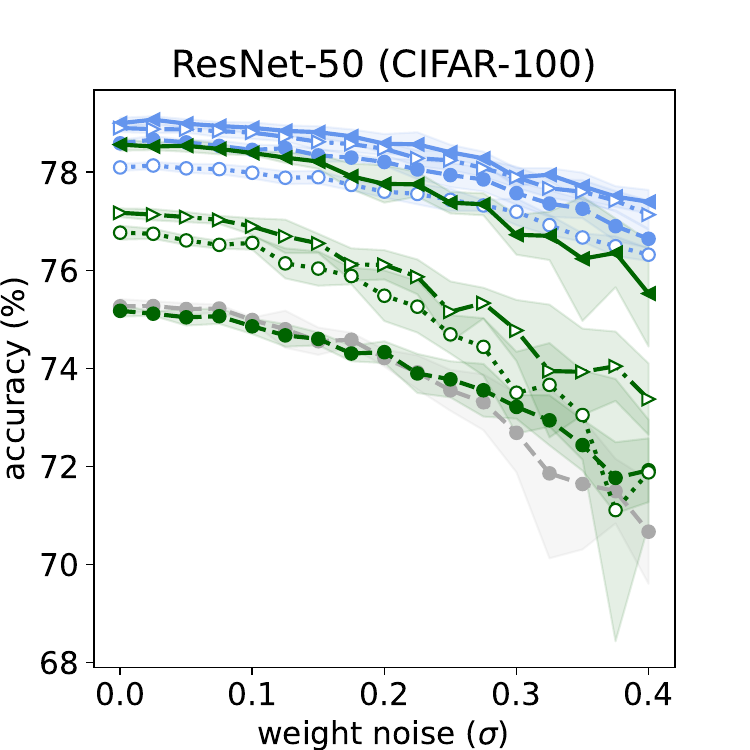}
     \end{subfigure}
     \hfill
     \begin{subfigure}[b]{0.32\textwidth}
         \centering
         \includegraphics[width=\textwidth]{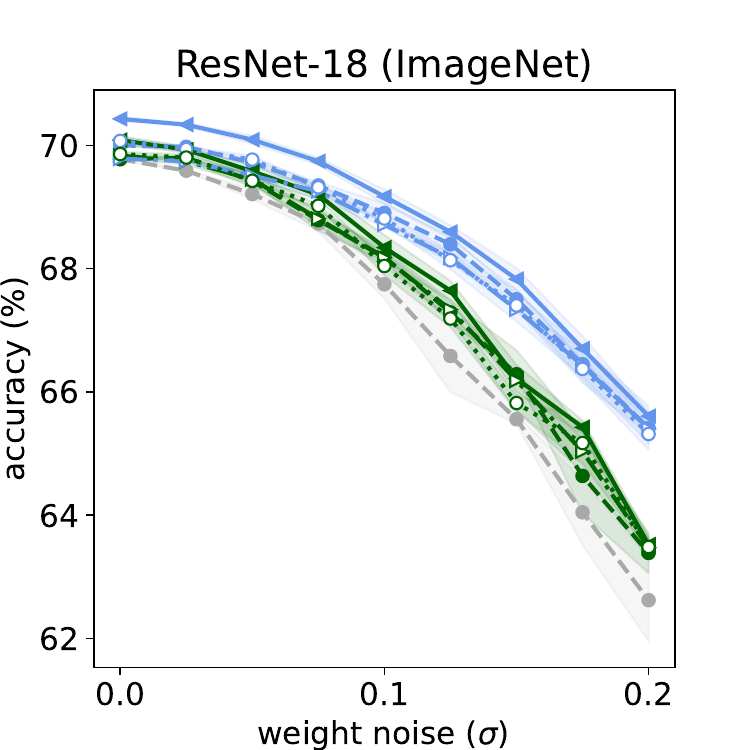}
     \end{subfigure}
     \begin{subfigure}[b]{0.6\textwidth}
         \centering
         \includegraphics[width=\textwidth]{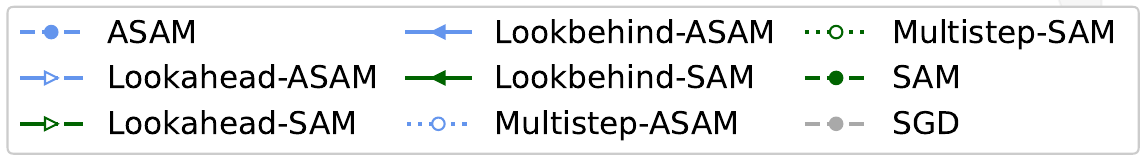}
     \end{subfigure}
     \hfill
        \caption{Robustness against noisy weights at inference time. We plot the mean and standard deviation over 10 and 3 inference runs for CIFAR-10/100 and ImageNet, respectively. Higher accuracy is better.
        }
        \label{fig:noisy_weights}
\end{figure*}

The sharpness over different radii of the different methods, when also trained with different $\rho$, are shown in Figure \ref{fig:sharpness_rho}. We observe that on top of Lookbehind improving sharpness at the nearby neighborhoods (as previously shown in Figure \ref{fig:loss_sharpness_trade_off}), models trained with Lookbehind also converge to flatter minima at the end of training, as measured on an extensive range of tested radii. This is consistent across training with different $\rho$ for both SAM and ASAM. Even though the minima found by the Lookahead and Multistep variants tend to have low sharpness with default $\rho$, such benefits diminish at higher $\rho$. 

\subsection{Model robustness}
\label{sec:robustness}

We now assess model robustness against noisy weights. This is a particularly important use case when deployment models in highly energy-efficient hardware implementations that are prone to variabilities and noise \citep{10.1145/2463209.2488867,kern2022memse,10.3389/fncom.2021.675741}. Similar to previous works \citep{joshi2020accurate,mordido2022sharpness}, we apply a multiplicative Gaussian noise to the model parameters $\phi$ after training in the form of $\phi \times \delta$, with $\delta \sim \mathcal{N}(1,\sigma^2)$ and update the
batch normalization statistics after the noise perturbations. Robustness results are presented in Figure \ref{fig:noisy_weights}. 

We see that Lookbehind shows the highest robustness observed by preserving the most amount of validation accuracy across the tested noise levels. This is observed for both SAM and ASAM on all models and datasets. We note that the benefits of using sharpness-aware minimization methods to increase model robustness to noisy weights were shown by previous works \citep{mordido2022sharpness}. Our results share these findings and further show that Lookbehind considerably boosts the robustness benefits of training with SAM and ASAM across several models and datasets.

\begin{table*}[t]
\caption{Lifelong learning performance in terms of average accuracy (higher is better) and forgetting (lower is better) on Split-CIFAR100 and Split-TinyImageNet.}
\begin{center}
\resizebox{0.8\textwidth}{!}{\begin{tabular}
{|l|cc|cc|}
\hline
Dataset & \multicolumn{2}{c|}{Split-CIFAR100}& \multicolumn{2}{c|}{Split-TinyImagenet}\\
Metric & Avg. accuracy $\uparrow$ & Forgetting $\downarrow$ & Avg. accuracy $\uparrow$ & Forgetting $\downarrow$\\
\hline\hline
SGD & $58.41_{\pm 4.95}$ & $22.74_{\pm 4.85}$ & $43.48_{\pm 0.80}$ & $26.51_{\pm 0.71}$\\
SAM & $57.81_{\pm 1.05}$ & $23.27_{\pm 0.57}$ & $56.34_{\pm 1.72}$ & $20.39_{\pm 1.83}$\\
{Multistep-SAM} & ${59.58_{\pm 0.34}}$ & ${15.09_{\pm 0.48}}$ & ${56.09_{\pm 1.17}}$ & ${20.70_{\pm 1.05}}$\\
\textbf{Lookbehind-SAM} & $\pmb{59.93_{\pm 1.54}}$ & $\pmb{14.10_{\pm 0.98}}$ & $\pmb{56.60_{\pm 0.68}}$ & $\pmb{18.99_{\pm 0.62}}$\\
\hline
ER + SGD & $64.84_{\pm 1.29}$ & $12.96_{\pm 0.23}$ & $49.19_{\pm 0.93}$ & $19.06_{\pm 0.26}$\\
ER + SAM & $68.28_{\pm 1.30}$ & $13.98_{\pm 0.42}$ & $65.59_{\pm 0.19}$ & $9.89_{\pm 0.14}$\\
ER + {Multistep-SAM} & ${65.49_{\pm 4.10}}$ & ${15.20_{\pm 2.53}}$ & ${65.75_{\pm 0.16}}$ & ${9.90_{\pm 0.09}}$\\
\textbf{ER + Lookbehind-SAM} & $\pmb{68.87_{\pm 0.79}}$ & $\pmb{12.37_{\pm 0.11}}$ & $\pmb{65.91_{\pm 0.27}}$ & $\pmb{9.11_{\pm 0.63}}$\\
\hline
Lookahead-C-MAML & $65.44_{\pm 0.99}$ & $13.96_{\pm 0.86}$ & $61.93_{\pm 1.55}$ & $11.53_{\pm 1.11}$\\
\textbf{Lookbehind-C-MAML} & ${\pmb{67.15_{\pm 0.74}}}$ & ${\pmb{12.40_{\pm 0.49}}}$ & ${\pmb{62.16_{\pm 0.86}}}$ & ${\pmb{11.21_{\pm 0.44}}}$\\
\hline
\end{tabular}}
\end{center}
\label{tab:lll}
\end{table*}

\begin{figure*}[h]
    \centering
    \begin{subfigure}[b]{1.\textwidth}
        \centering
        \includegraphics[width=1.\linewidth]{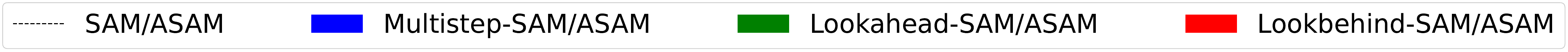}
    \end{subfigure}
    \begin{subfigure}[b]{0.49\textwidth}
        \centering
        \includegraphics[width=0.475\linewidth]{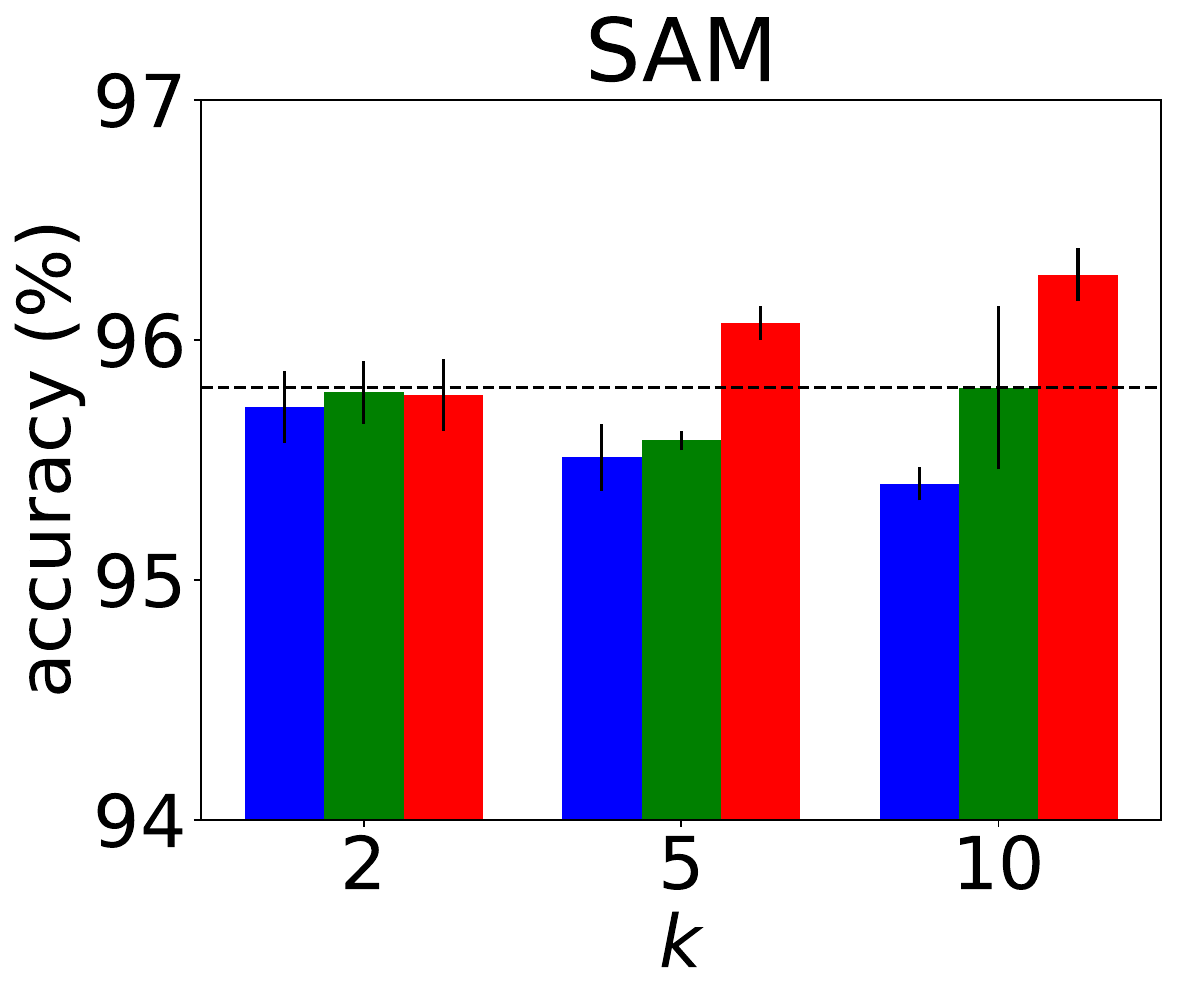}
        \hfill
        \includegraphics[width=0.475\linewidth]{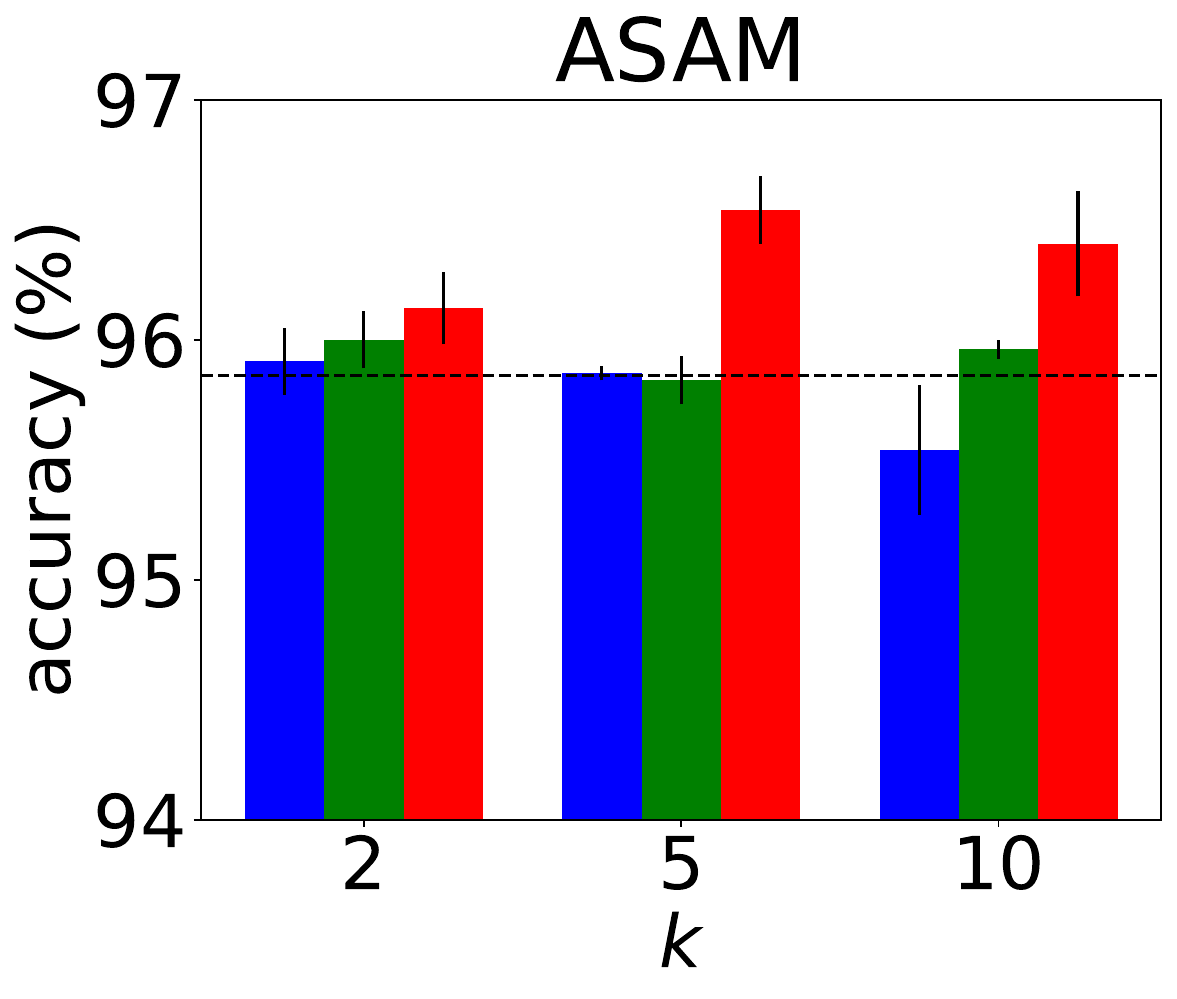}
        \caption{ResNet-34 on CIFAR-10.}
    \end{subfigure}
    \hfill
    \begin{subfigure}[b]{0.49\textwidth}
        \centering
        \includegraphics[width=0.475\linewidth]{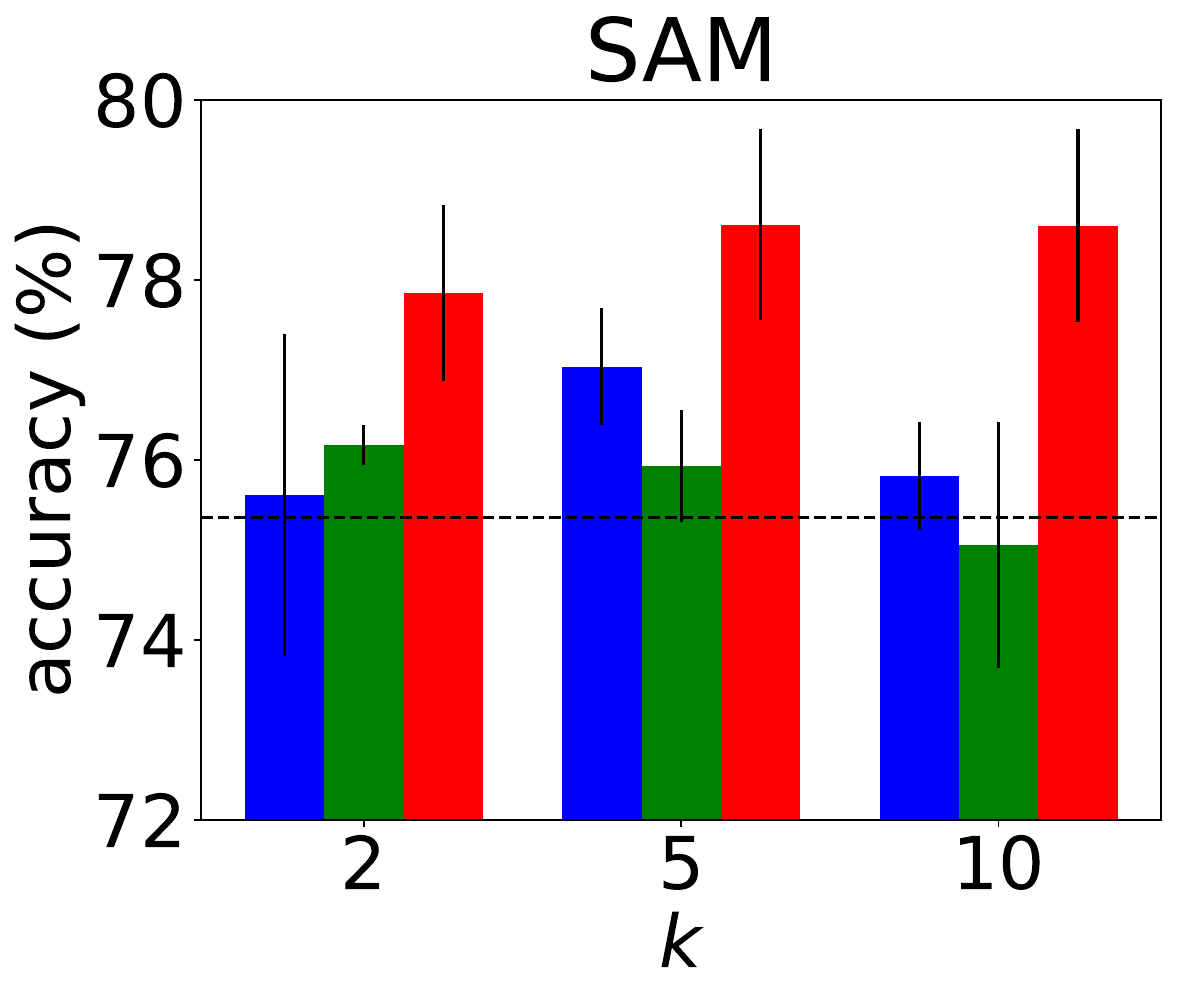}
        \hfill
        \includegraphics[width=0.475\linewidth]{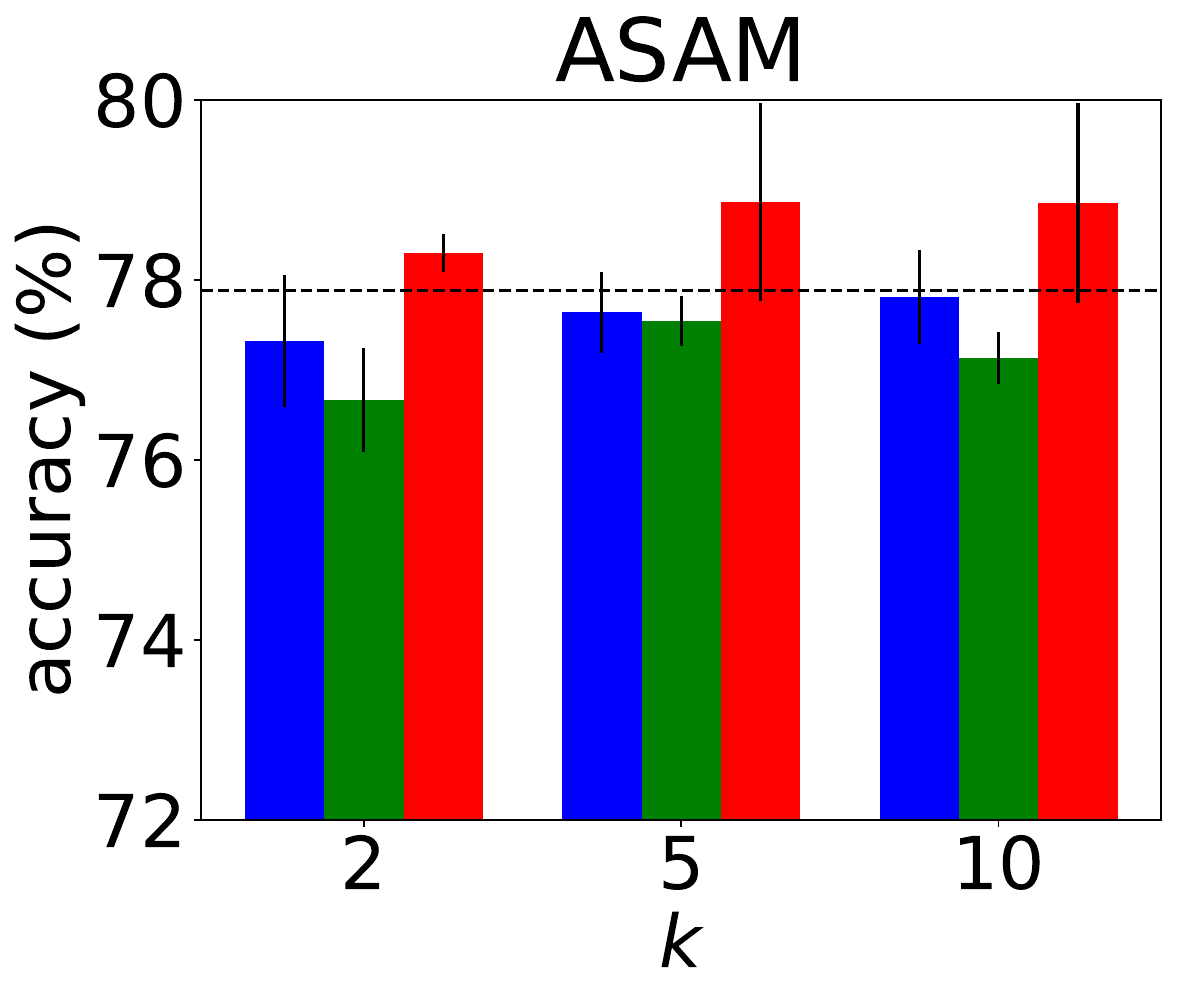}
        \caption{ResNet-50 on CIFAR-100.}
    \end{subfigure}
    \caption{Generalization performance (validation acc. \%) between Multistep-SAM/SAM, Lookahead-SAM/ASAM, and Lookbehind-SAM/ASAM. Vanilla SAM and ASAM baselines with default $\rho$ are represented by the horizontal line.
    }
    \label{fig:ablation}
\end{figure*}

\subsection{Lifelong learning}
\label{sec:lifelong_learning}

Lastly, we evaluate the methods in lifelong learning where a model with a limited capacity is trained on a stream of tasks. The goal is then to maximize performance across tasks without having access to previous data.  
In our experiments, we replicate the same setup used in Lookahead-MAML \citep{gupta2020look}, which is a lifelong learning method that combines the concept of slow and fast weights of Lookahead with meta-learning principles \citep{finn2017model}. Moreover, we replace Lookahead with Lookbehind, creating a novel algorithm: Lookbehind-MAML. Since meta-learning is out of the scope of this work, we implemented only the constant learning rate setting for simplicity, \textit{i.e.} the C-MAML variant \citep{gupta2020look}. 

We train a 3- and a 4-layer convolutional network on Split-CIFAR100 and Split-TinyImageNet, respectively. We report the following metrics by evaluating the model on the held-out data set: average accuracy (higher is better) and forgetting (lower is better).
Additional details about the algorithms, training, and datasets are provided in Appendix \ref{sec:app_lll_exp_details}. 
The results are presented in Table \ref{tab:lll}. In the first setting, we do not use ER and directly compare our method with SGD, SAM, and Multistep-SAM. We observe that Lookbehind achieves the best performance both in terms of average accuracy and forgetting. In the second setting, we apply ER to the previous methods. Once again, we see an improvement when using our variant. Finally, when comparing Lookahead-C-MAML with Lookbehind-C-MAML, we also notice an overall performance improvement.

\section{Sensitivity analysis}

In this section, we analyze the sensitivity of Lookbehind to different hyper-parameter settings in terms of generalization performance (Sections \ref{sec:ablations}, \ref{sec:sensitivity_rho}, and \ref{sec:sensitivity_alpha}) and its benefits at different training stages (Section \ref{sec:training_stages}). 
For the following experiments, we used ResNet-34 and ResNet-50 models trained from scratch on CIFAR-10 and CIFAR-100, respectively. Training and hyperparameter search details are provided in Appendices \ref{sec:app_training_details} and \ref{sec:hyperparameter_tuning}.

\subsection{Sensitivity to the inner step $k$}
\label{sec:ablations}

\begin{figure*}[t]
     \centering
     \begin{subfigure}[b]{0.32\textwidth}
         \centering
         \includegraphics[width=\textwidth]{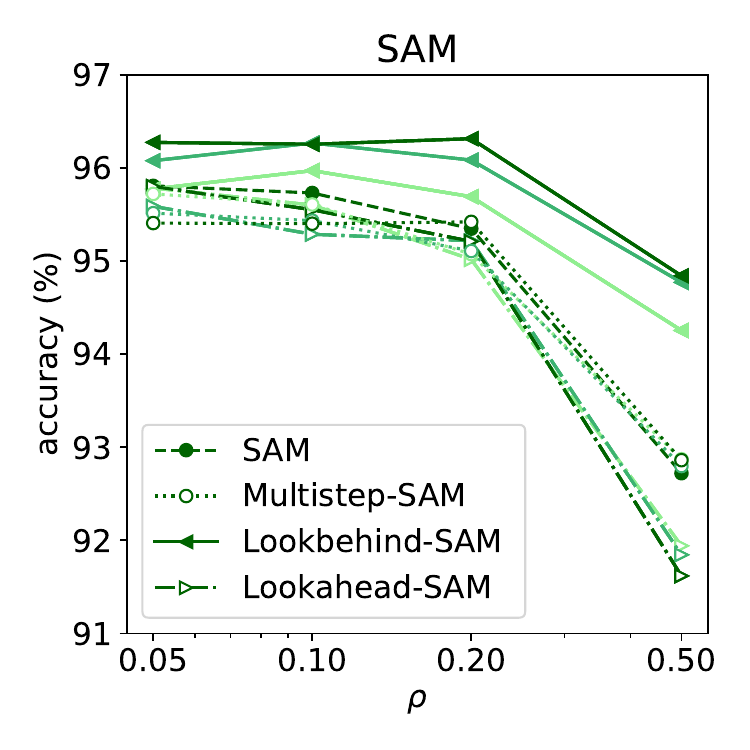}
     \end{subfigure}
     \begin{subfigure}[b]{0.32\textwidth}
         \centering
         \includegraphics[width=\textwidth]{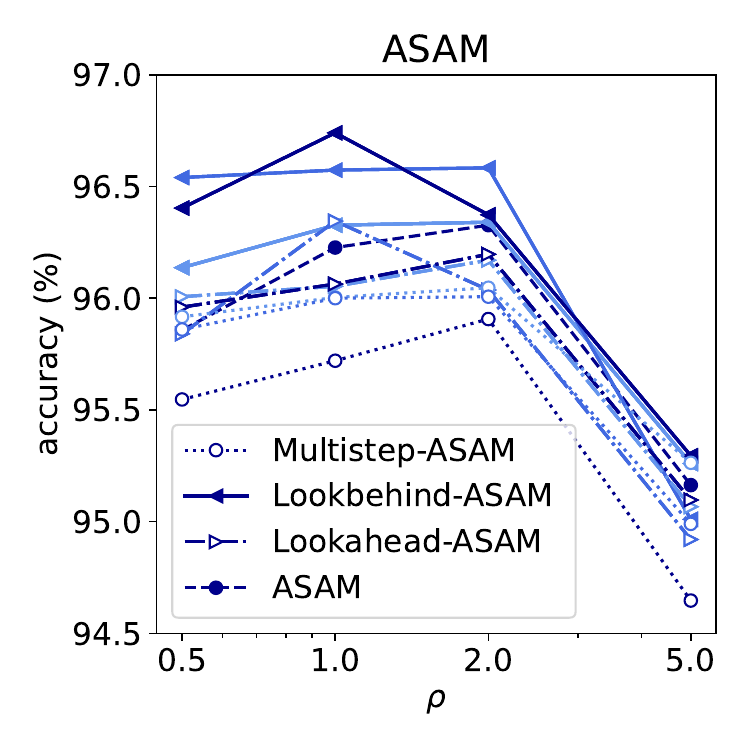}
     \end{subfigure}
     \hfill
        \caption{Validation accuracies with different trained $\rho$ for the different methods using ResNet-34 trained on CIFAR-10. Darker shades represent larger inner steps $k$, ranging from $k \in \{2, 5, 10\}$. Higher accuracy is better.}
        \label{fig:rho_test_accuracies}
\end{figure*}

Validation accuracies of the different methods when using different $k$ are presented in Figure \ref{fig:ablation}. We observe that Lookbehind is the only method that consistently outperforms the SAM and ASAM baselines on both CIFAR-10 and CIFAR-100, across all the tested inner steps $k$. Interestingly, our method tends to keep improving when increasing $k$, while this trend is not observed for either the Lookahead or the Multistep variants.
Moreover, we see that Multistep-SAM/ASAM does not provide a clear improvement over the respective SAM and ASAM baselines, as previously discussed in prior work \citep{foret2021sharpnessaware,andriushchenko2022towards}.
On the other hand, the Lookahead variants show a slight improvement over Multistep, particularly when combining Lookahead with SAM and ASAM on CIFAR-10 and SAM on CIFAR-100. Overall, we see that Lookbehind is the highest-performing method on all models and datasets when combined with SAM/ASAM.

\subsection{Sensitivity to the neighborhood size $\rho$}
\label{sec:sensitivity_rho}

We now analyze the effects of training with increasing $\rho$ with the different methods using SAM and ASAM. 
Results are presented in Figure \ref{fig:rho_test_accuracies}. We see that our method is the only one that consistently outperforms SAM and ASAM across all the tested $\rho$. As previously suggested, significantly increasing $\rho$ in SAM, \textit{e.g.} $\rho = 0.5$, decreases performance relative to the default $\rho$, \textit{i.e.} $\rho = 0.05$. Similarly, increasing $\rho$ in ASAM also decreases performance relative to its default $\rho$ of $0.5$. Notwithstanding, we note that ASAM shows higher relative robustness to higher $\rho$ than SAM, indicated by ASAM's ability to continue increasing performance on up to 4$\times$ the default neighborhood size, \textit{i.e.} from $\rho=0.5$ to $\rho=2.0$. Overall, we observe that Lookbehind is more robust to the choice of $\rho$ compared to the other methods, with Lookbehind and Multistep variants showing similar trends as the SAM and ASAM baselines. 

\subsection{Sensitivity to the outer step size $\alpha$}
\label{sec:sensitivity_alpha}

\begin{figure}[t]
     \centering
     \begin{subfigure}[b]{0.49\textwidth}
         \centering
         \includegraphics[trim=0 260.0pt 0 0, clip, width=0.49\textwidth]{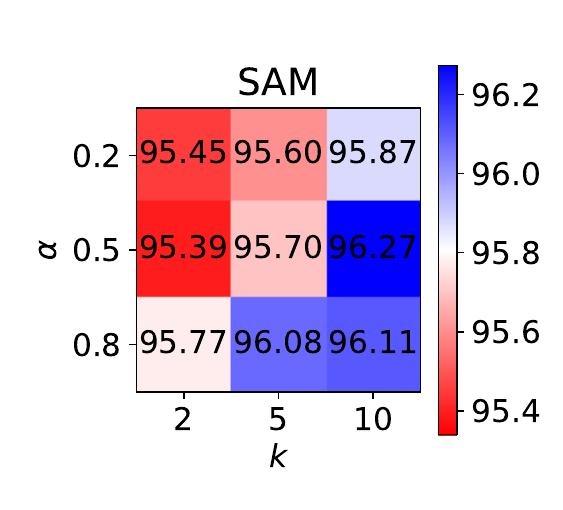}
         \hfill
         \includegraphics[trim=0 15pt 0 260.0pt, clip, width=0.49\textwidth]{Figures/test_accuracies_resnet34.pdf}
      \end{subfigure}
        \caption{Sensitivity of Lookbehind to $\alpha$ and $k$ using ResNet-34 on CIFAR-10 in terms of validation accuracy (\%). The vanilla SAM/ASAM performances are in white.
        }
        \label{fig:lookafar_alpha}
\end{figure}

The validation accuracies of Lookbehind across different $\alpha$ and $k$ are presented in Figure \ref{fig:lookafar_alpha}. All models were trained with the default $\rho$, with blue representing an improvement over SAM/ASAM and red a degradation. We see that Lookbehind improves over the baselines in all $k$, except $k=2$ on SAM and CIFAR-10. 
We notice a diagonal trend, suggesting there is a relation between $\alpha$ and $k$, with a larger $\alpha$ being often better with smaller $k$ and a mid to high range $\alpha$ working well with higher $k$. 
An in-depth analysis across all models and datasets is provided in Appendix \ref{sec:app_sensitivity}, showcasing that Lookbehind is generally robust to specific combinations of $k$ and $\alpha$ across different ranges.

\begin{figure*}[t]
     \centering
     \begin{subfigure}[b]{0.49\textwidth}
         \centering
         \includegraphics[width=0.49\textwidth]{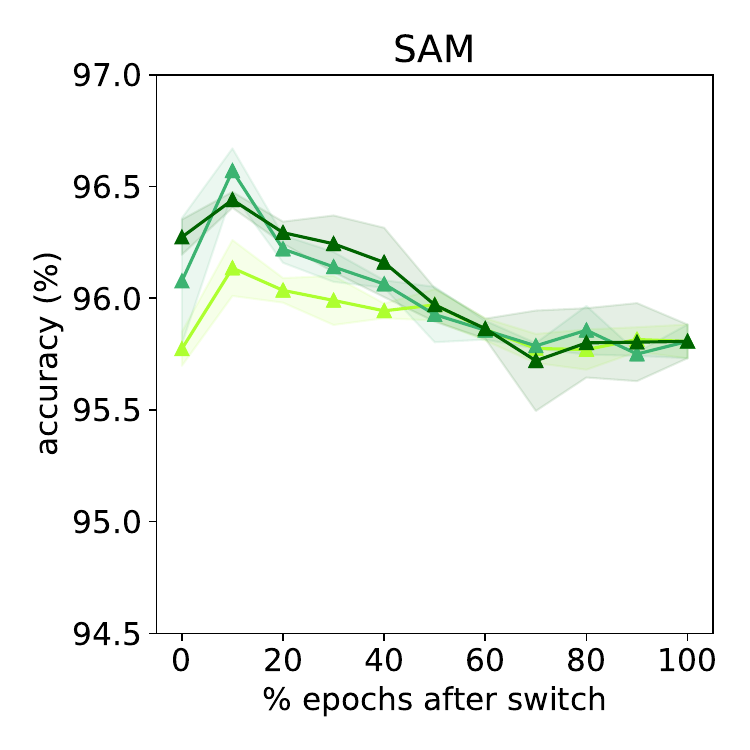}
         \hfill
         \includegraphics[width=0.49\textwidth]{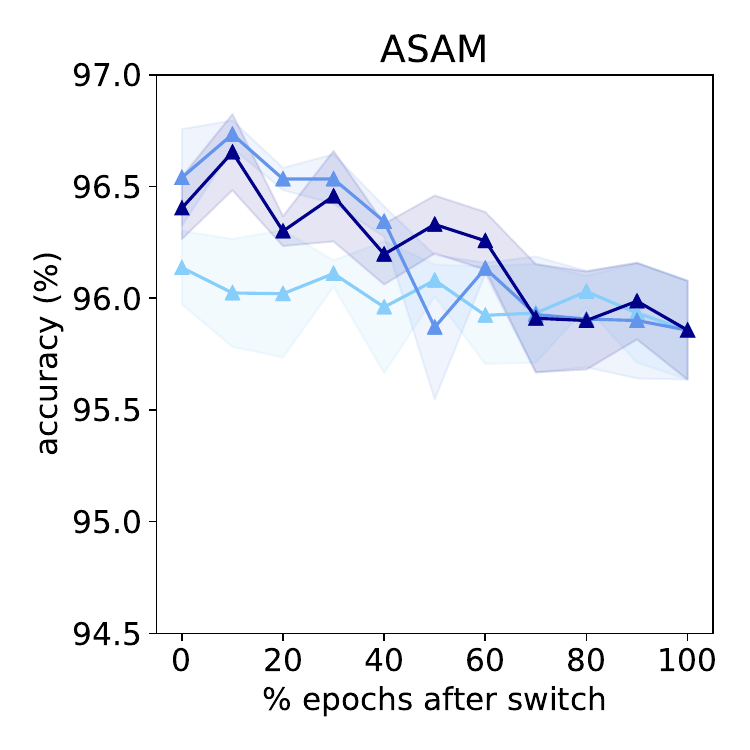}
         \caption{SAM/ASAM $\rightarrow$ Lookbehind-SAM/ASAM}
         \label{fig:sam_lookbehind}
     \end{subfigure}
     \hfill
     \begin{subfigure}[b]{0.49\textwidth}
         \centering
         \includegraphics[width=0.49\textwidth]{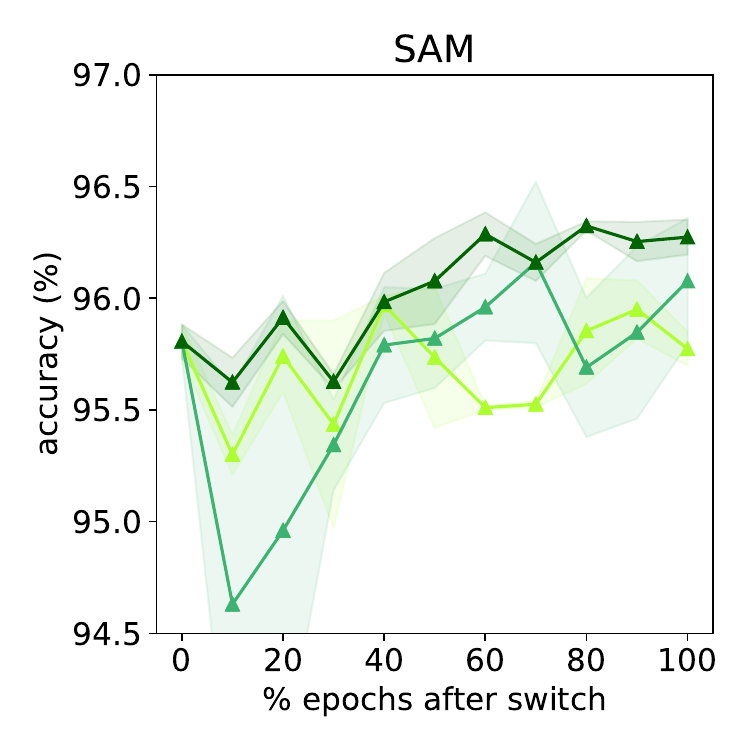}
         \hfill
         \includegraphics[width=0.49\textwidth]{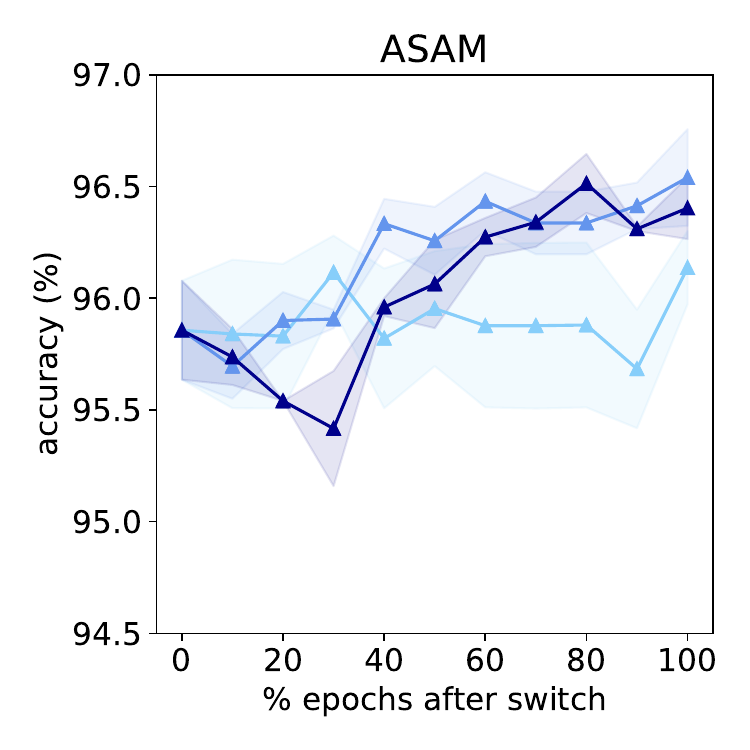}
         \caption{Lookbehind-SAM/ASAM $\rightarrow$ SAM/ASAM}
         \label{fig:lookbehind_sam}
     \end{subfigure}
        \caption{Impact of switching from SAM/ASAM to Lookbehind-SAM/ASAM (a), and vice-versa (b), at different epochs throughout training in terms of validation accuracy using ResNet-34 trained on CIFAR-10. Darker shades represent larger inner steps $k$, ranging from $k \in \{2, 5, 10\}$. For Lookbehind, we pick the best $\alpha$ configuration for each $k \in \{2, 5, 10\}$ using the default $\rho$, which is also used for the SAM/ASAM baselines.
        }
        \label{fig:switch_test_accuracies}
\end{figure*}

\begin{table*}[t]
\caption{Generalization performance (validation acc. \%) of Lookbehind with static and adaptive $\alpha$.}
\begin{center}
\resizebox{0.8\textwidth}{!}{
\begin{tabular}{|l|c|c||c|c||c|}
\hline
Dataset & \multicolumn{2}{c||}{CIFAR-10} & \multicolumn{2}{c||}{CIFAR-100} & ImageNet\\
Model & ResNet-34 & WRN-28-2 & ResNet-50 & WRN-28-10 & ResNet-18\\
\hline\hline
Lookbehind-SAM & $96.27_{\pm .07}$  & $94.81_{\pm .22}$  & $\pmb{78.62_{\pm .48}}$ & $\pmb{80.99_{\pm .02}}$ & $\pmb{70.16_{\pm .08}}$\\
\quad + adaptive $\alpha$ & $\pmb{96.33_{\pm .04}}$  & $\pmb{94.88_{\pm .12}}$  & $78.33_{\pm .36}$ & $80.86_{\pm .13}$ & $70.07_{\pm .12}$\\
\hline
Lookbehind-ASAM & $96.54_{\pm .21}$  & $\pmb{95.23_{\pm .01}}$  & $78.86_{\pm .29}$ & $\pmb{82.16_{\pm .09}}$ & $\pmb{70.23_{\pm .22}}$\\
\quad + adaptive $\alpha$ & $\pmb{96.57_{\pm .03}}$ & $95.08_{\pm .15}$  & $\pmb{78.89_{\pm .45}}$ & $81.86_{\pm .22}$& $70.16_{\pm .08}$\\
\hline
\end{tabular}}
\end{center}
\label{tab:adaptive_alpha}
\end{table*}

\subsection{Benefits of Lookbehind at different training stages}
\label{sec:training_stages}

SAM has been shown to find better generalizable minima within the same basin as SGD.
In other words, SAM's implicit bias mostly improves the generalization of SGD when switching from SGD to SAM toward the end of training \citep{andriushchenko2022towards}.
Interestingly, the aforementioned results also suggest that SAM and SGD do not guide optimization toward different basins from early on in training. Here, we conduct a similar study by analyzing how switching from SAM/ASAM to Lookbehind-SAM/ASAM, and vice-versa, impacts generalization performance at different stages during training.

The generalization performances of starting training with SAM/ASAM and switching to Lookbehind at different training stages are shown in Figure \ref{fig:sam_lookbehind}. We observe that Lookbehind's benefits are mostly achieved early on in training, suggesting that Lookbehind guides the optimization to converge to a different basin of the loss landscape than SAM. Such findings are confirmed by also switching from Lookbehind to SAM/ASAM (Figure \ref{fig:lookbehind_sam}).

\section{Discussions and limitations}

One limitation of Lookbehind is that it adds two additional hyperparameters to SAM/ASAM – just as the Lookahead optimizer adds two hyperparameters to SGD. This introduces additional hyperparameter tuning on top of $\rho$ and $\eta$. To alleviate this concern in settings where computational resources are scarce, we experiment with removing the need to tune $\alpha$ by computing it analytically during training (Section \ref{sec:adaptive_alpha}).

Another important limitation inherent to any multiple ascent step SAM method is the computational overhead which increases training time by a factor $k$. Even though the goal of this work is to tackle the lack of performance due to a poor sharpness-loss trade-off, this limitation may prevent training with Lookbehind on larger models. To address this issue, we explore improving the efficiency of multiple ascent steps by switching the minibatch at each inner step of Lookbehind (Section \ref{sec:switch_minibatch}).

\subsection{Adaptive $\alpha$}
\label{sec:adaptive_alpha}

Here, we propose an adaptive formulation of $\alpha$, defined as $\alpha^*$, by setting it proportionally to the alignment of the gradients obtained during the multiple ascent steps:
\begin{equation}
    \alpha^* = (\cos{(\theta)}+1)/2 \,,
\end{equation}
where $\theta$ is defined by the angle between the first gathered gradient and the final update direction:
\begin{equation}
    \theta = \dfrac{(\phi_{t,1}-\phi_t) \cdot (\phi_{t,k}-\phi_t)}{\|\phi_{t,1}-\phi_t\|_2 \cdot \|\phi_{t,k}-\phi_t\|_2} \,.
\end{equation}
If the gradients are completely aligned, then $\alpha^* = 1$. Alternatively, if the gradients are not aligned, then $0 {\,\,\leq \,\,} \alpha^* < 1$, with lower values representing lower alignment.

Results when using Lookbehind with a static $\alpha$ and a dynamic $\alpha^*$ are presented in Table \ref{tab:adaptive_alpha}. Overall, we observe that using an adaptive $\alpha$ is a viable alternative to tuning a static $\alpha$ in instances where compute is scarce. Note that our goal with adaptive $\alpha$ is not necessarily to outperform static $\alpha$ but instead to achieve competitive performance while having one less hyperparameter. Importantly, we emphasize that Lookbehind with adaptive $\alpha$ consistently outperforms all the compared methods presented in Table \ref{tab:generalization}, similarly to static $\alpha$. We refer to Appendix \ref{sec:additional_discussions} for additional discussions and an analysis of how $\alpha^*$ varies during training.

\subsection{Switching minibatch}
\label{sec:switch_minibatch}

Performing multiple ascent steps with the same minibatch reduces the variance of the aggregated gradients. However, this also increases training time due to redundant gradient computations. In time-sensitive scenarios like training Transformers, switching the minibatch during ascent steps might be the only viable option. This simple modification involves moving minibatch sampling (line 4, Algorithm \ref{alg:lookbehind}) inside the inner loop in our method.

We consider machine translation by replicating the setup in \citet{kwon2021asam} and using a 12-layer Transformer \cite{vaswani2017attention} (39.4M parameters) trained from scratch on IWSLT’14 (DE-EN) with Adam. We also perform image classification by finetuning a ViT-Base model \citep{dosovitskiy2021an} (86.6M parameters) for 15 epochs with SGD. The model was previously pre-trained on ImageNet-21K and finetuned with SGD on ImageNet-1K. Appendices \ref{sec:app_training_details} and \ref{sec:hyperparameter_tuning} provide more details.

Results under a similar training budget as SAM/ASAM by switching the minibatch in Lookbehind are presented in Table \ref{tab:transformers}. We observe that at least one Lookbehind variant (with a static or dynamic $\alpha$) improves the performance of SAM/ASAM. We note that in the two instances where a specific Lookbehind variant was unable to outperform the ASAM baseline, their performance was almost identical.
We highlight that these settings showcase the applicability of Lookbehind to additional optimizers (\textit{i.e.} Adam) and settings (\textit{i.e.} fine-tuning of large models).

\begin{table}[h]
\caption{Generalization performance with Transformers on machine translation (test BLEU score) and image classification (validation acc. \%). We switch the minibatch for Lookbehind and use $k=2$ with $\alpha=0.8$ or adaptive $\alpha$.
}
\begin{center}
\resizebox{0.9\linewidth}{!}{
\begin{tabular}{|l|c||c|}
\hline
Dataset & IWSLT’14 & ImageNet\\
Model & Transformer & ViT-Base\\
\hline\hline
Adam/SGD & $34.86_{\pm .01}$ & $81.79$\\
\hline
SAM & $34.78_{\pm .01}$ & $81.80$\\
Lookbehind-SAM & $\pmb{35.10_{\pm .01}}$ & $\pmb{81.84}$\\
\quad + adaptive $\alpha$ & $\pmb{35.22_{\pm .01}}$  & $\pmb{81.85}$\\
\hline
ASAM & $35.02_{\pm .01}$  & $\pmb{81.89}$\\
Lookbehind-ASAM & $\pmb{35.40_{\pm .01}}$  & $81.87$\\
\quad + adaptive $\alpha$ & $35.00_{\pm .01}$ & $\pmb{81.89}$\\
\hline
\end{tabular}}
\end{center}
\label{tab:transformers}
\end{table}

\section{Related work}

Sharpness-aware minimization (SAM) \citep{foret2021sharpnessaware} is an attempt to improve generalization by finding solutions with both low loss value and low loss sharpness. 
This is achieved by minimizing an estimation of the maximum loss over 
a neighborhood region around the parameters. 
There is currently a lot of active work that focuses on improving SAM. More specifically, modifications of the original SAM algorithm were proposed to further improve generalization performance \citep{zhuang2022surrogate,kim2022fisher,kwon2021asam,liu2022random} and efficiency \citep{du2022sharpnessaware,zhou2021delta,liu2022towards}. Performing multiple ascent steps was present in \citet{foret2021sharpnessaware}, however, the improvements over single ascent step SAM were either insignificant or shown to degrade performance in some settings \citep{andriushchenko2022towards}.

SAM's benefits have transcended improving generalization performance, ranging from higher robustness to label noise \citep{foret2021sharpnessaware,kwon2021asam,baek2024why}, lower quantization error
\citep{liu2021sharpness}, and less sensitivity to data imbalance \citep{liu2021selfsupervised}. Here, on top of analyzing the benefits of Lookbehind in terms of generalization, we build on the recently observed benefits of SAM on improving robustness against noisy weights \citep{kim2022fisher,mordido2022sharpness} and reducing catastrophic forgetting in lifelong learning \citep{mehta2021empirical}.

Closest to our work, \citet{kim2023exploring} concurrently conducted a similar study by averaging the gradients obtained during multiple SAM ascent steps.
One of the differences is the decoupling of the inner step $k$ and the outer step size $\alpha$ in our approach, which allows us to seek optimal combinations between these two hyperparameters. As depicted in Figures \ref{fig:lookafar_alpha} and \ref{fig:lookbehind_app}, $\alpha = 1/k$ is generally not the best overall $\alpha$ to use, including when determining $\alpha^*$ (Figure \ref{fig:adaptive_alpha_evolution}).
We also extend the empirical discussions by applying our method with ASAM, which often outperforms SAM.

\section{Conclusion}

In this work, we proposed Lookbehind, which can be plugged on top of existing sharpness-aware training methods to improve model performance across a variety of tasks and benchmarks. More specifically, we show an improvement in generalization performance on multiple models and datasets, model robustness, and continuous learning ability. Moreover, we propose two simple method modifications to address limitations that may arise in large-scale settings, broadening the applicability of our approach. 
In the future, exploring novel approaches or combining Lookbehind with more efficient SAM variants to mitigate the computational overhead of multiple ascent step SAM is worth pursuing.

\section*{Acknowledgements}

Gonçalo Mordido was supported by an FRQNT postdoctoral scholarship (PBEEE) during part of this work. Sarath Chandar is supported by the Canada CIFAR AI Chairs program, the Canada Research Chair in Lifelong Machine Learning, and the NSERC Discovery Grant. This research was enabled in part by compute resources provided by Mila (mila.quebec).

\section*{Impact Statement}

This paper presents work whose goal is to advance the field of Machine Learning. There are many potential societal consequences of our work, none of which we feel must be specifically highlighted here.

\bibliography{example_paper}

\begin{thebibliography}{39}
\providecommand{\natexlab}[1]{#1}
\providecommand{\url}[1]{\texttt{#1}}
\expandafter\ifx\csname urlstyle\endcsname\relax
  \providecommand{\doi}[1]{doi: #1}\else
  \providecommand{\doi}{doi: \begingroup \urlstyle{rm}\Url}\fi

\bibitem[Andriushchenko \& Flammarion(2022)Andriushchenko and Flammarion]{andriushchenko2022towards}
Andriushchenko, M. and Flammarion, N.
\newblock Towards understanding sharpness-aware minimization.
\newblock In \emph{International Conference on Machine Learning}, 2022.

\bibitem[Baek et~al.(2024)Baek, Kolter, and Raghunathan]{baek2024why}
Baek, C., Kolter, J.~Z., and Raghunathan, A.
\newblock Why is {SAM} robust to label noise?
\newblock In \emph{International Conference on Learning Representations}, 2024.

\bibitem[Chaudhry et~al.(2018)Chaudhry, Dokania, Ajanthan, and Torr]{chaudhry2018riemannian}
Chaudhry, A., Dokania, P.~K., Ajanthan, T., and Torr, P.~H.
\newblock Riemannian walk for incremental learning: Understanding forgetting and intransigence.
\newblock In \emph{European Conference on Computer Vision}, 2018.

\bibitem[Chaudhry et~al.(2019)Chaudhry, Rohrbach, Elhoseiny, Ajanthan, Dokania, Torr, and Ranzato]{chaudhry2019tiny}
Chaudhry, A., Rohrbach, M., Elhoseiny, M., Ajanthan, T., Dokania, P.~K., Torr, P.~H., and Ranzato, M.
\newblock On tiny episodic memories in continual learning.
\newblock \emph{arXiv preprint arXiv:1902.10486}, 2019.

\bibitem[Deng et~al.(2009)Deng, Dong, Socher, Li, Li, and Fei-Fei]{deng2009imagenet}
Deng, J., Dong, W., Socher, R., Li, L.-J., Li, K., and Fei-Fei, L.
\newblock {ImageNet}: A large-scale hierarchical image database.
\newblock In \emph{IEEE Conference on Computer Vision and Pattern Recognition}, 2009.

\bibitem[Dinh et~al.(2017)Dinh, Pascanu, Bengio, and Bengio]{dinh2017sharp}
Dinh, L., Pascanu, R., Bengio, S., and Bengio, Y.
\newblock Sharp minima can generalize for deep nets.
\newblock In \emph{International Conference on Machine Learning}, 2017.

\bibitem[Dosovitskiy et~al.(2021)Dosovitskiy, Beyer, Kolesnikov, Weissenborn, Zhai, Unterthiner, Dehghani, Minderer, Heigold, Gelly, Uszkoreit, and Houlsby]{dosovitskiy2021an}
Dosovitskiy, A., Beyer, L., Kolesnikov, A., Weissenborn, D., Zhai, X., Unterthiner, T., Dehghani, M., Minderer, M., Heigold, G., Gelly, S., Uszkoreit, J., and Houlsby, N.
\newblock An image is worth 16x16 words: Transformers for image recognition at scale.
\newblock In \emph{International Conference on Learning Representations}, 2021.

\bibitem[Du et~al.(2022{\natexlab{a}})Du, Daquan, Feng, Tan, and Zhou]{du2022sharpnessaware}
Du, J., Daquan, Z., Feng, J., Tan, V., and Zhou, J.~T.
\newblock Sharpness-aware training for free.
\newblock In \emph{Advances in Neural Information Processing Systems}, 2022{\natexlab{a}}.

\bibitem[Du et~al.(2022{\natexlab{b}})Du, Yan, Feng, Zhou, Zhen, Goh, and Tan]{du2022efficient}
Du, J., Yan, H., Feng, J., Zhou, J.~T., Zhen, L., Goh, R. S.~M., and Tan, V.
\newblock Efficient sharpness-aware minimization for improved training of neural networks.
\newblock In \emph{International Conference on Learning Representations}, 2022{\natexlab{b}}.

\bibitem[Dziugaite \& Roy(2017)Dziugaite and Roy]{DR17}
Dziugaite, G.~K. and Roy, D.~M.
\newblock Computing nonvacuous generalization bounds for deep (stochastic) neural networks with many more parameters than training data.
\newblock In \emph{Conference on Uncertainty in Artificial Intelligence}, 2017.

\bibitem[Finn et~al.(2017)Finn, Abbeel, and Levine]{finn2017model}
Finn, C., Abbeel, P., and Levine, S.
\newblock Model-agnostic meta-learning for fast adaptation of deep networks.
\newblock In \emph{International Conference on Machine Learning}, 2017.

\bibitem[Foret et~al.(2021)Foret, Kleiner, Mobahi, and Neyshabur]{foret2021sharpnessaware}
Foret, P., Kleiner, A., Mobahi, H., and Neyshabur, B.
\newblock Sharpness-aware minimization for efficiently improving generalization.
\newblock In \emph{International Conference on Learning Representations}, 2021.

\bibitem[Gupta et~al.(2020)Gupta, Yadav, and Paull]{gupta2020look}
Gupta, G., Yadav, K., and Paull, L.
\newblock Look-ahead meta learning for continual learning.
\newblock \emph{Advances in Neural Information Processing Systems}, 2020.

\bibitem[He et~al.(2016)He, Zhang, Ren, and Sun]{he2016deep}
He, K., Zhang, X., Ren, S., and Sun, J.
\newblock Deep residual learning for image recognition.
\newblock In \emph{IEEE Conference on Computer Vision and Pattern Recognition}, 2016.

\bibitem[Hochreiter \& Schmidhuber(1994)Hochreiter and Schmidhuber]{hochreiter1994simplifying}
Hochreiter, S. and Schmidhuber, J.
\newblock Simplifying neural nets by discovering flat minima.
\newblock \emph{Advances in Neural Information Processing Systems}, 1994.

\bibitem[Izmailov et~al.(2018)Izmailov, Podoprikhin, Garipov, Vetrov, and Wilson]{DBLP:conf/uai/IzmailovPGVW18}
Izmailov, P., Podoprikhin, D., Garipov, T., Vetrov, D.~P., and Wilson, A.~G.
\newblock Averaging weights leads to wider optima and better generalization.
\newblock In \emph{Conference on Uncertainty in Artificial Intelligence}, 2018.

\bibitem[Joshi et~al.(2020)Joshi, Le~Gallo, Haefeli, Boybat, Nandakumar, Piveteau, Dazzi, Rajendran, Sebastian, and Eleftheriou]{joshi2020accurate}
Joshi, V., Le~Gallo, M., Haefeli, S., Boybat, I., Nandakumar, S.~R., Piveteau, C., Dazzi, M., Rajendran, B., Sebastian, A., and Eleftheriou, E.
\newblock Accurate deep neural network inference using computational phase-change memory.
\newblock \emph{Nature Communications}, 2020.

\bibitem[Kern et~al.(2022)Kern, Henwood, Mordido, Dupraz, A{\"\i}ssa-El-Bey, Savaria, and Leduc-Primeau]{kern2022memse}
Kern, J., Henwood, S., Mordido, G., Dupraz, E., A{\"\i}ssa-El-Bey, A., Savaria, Y., and Leduc-Primeau, F.
\newblock {MemSE}: Fast {MSE} prediction for noisy memristor-based {DNN} accelerators.
\newblock In \emph{IEEE International Conference on Artificial Intelligence Circuits and Systems}, 2022.

\bibitem[Keskar et~al.(2016)Keskar, Mudigere, Nocedal, Smelyanskiy, and Tang]{keskar2016large}
Keskar, N.~S., Mudigere, D., Nocedal, J., Smelyanskiy, M., and Tang, P. T.~P.
\newblock On large-batch training for deep learning: Generalization gap and sharp minima.
\newblock In \emph{International Conference on Learning Representations}, 2016.

\bibitem[Kim et~al.(2023)Kim, Park, Choi, Lee, and Lee]{kim2023exploring}
Kim, H., Park, J., Choi, Y., Lee, W., and Lee, J.
\newblock Exploring the effect of multi-step ascent in sharpness-aware minimization.
\newblock \emph{arXiv preprint arXiv:2302.10181}, 2023.

\bibitem[Kim et~al.(2022)Kim, Li, Hu, and Hospedales]{kim2022fisher}
Kim, M., Li, D., Hu, S.~X., and Hospedales, T.
\newblock {Fisher SAM}: Information geometry and sharpness aware minimisation.
\newblock In \emph{International Conference on Machine Learning}, 2022.

\bibitem[Krizhevsky et~al.(2009)Krizhevsky, Hinton, et~al.]{krizhevsky2009learning}
Krizhevsky, A., Hinton, G., et~al.
\newblock Learning multiple layers of features from tiny images.
\newblock 2009.

\bibitem[Kwon et~al.(2021)Kwon, Kim, Park, and Choi]{kwon2021asam}
Kwon, J., Kim, J., Park, H., and Choi, I.~K.
\newblock {ASAM}: Adaptive sharpness-aware minimization for scale-invariant learning of deep neural networks.
\newblock In \emph{International Conference on Machine Learning}, 2021.

\bibitem[Liu et~al.(2021{\natexlab{a}})Liu, HaoChen, Gaidon, and Ma]{liu2021selfsupervised}
Liu, H., HaoChen, J.~Z., Gaidon, A., and Ma, T.
\newblock Self-supervised learning is more robust to dataset imbalance.
\newblock In \emph{NeurIPS 2021 Workshop on Distribution Shifts: Connecting Methods and Applications}, 2021{\natexlab{a}}.

\bibitem[Liu et~al.(2021{\natexlab{b}})Liu, Cai, and Zhuang]{liu2021sharpness}
Liu, J., Cai, J., and Zhuang, B.
\newblock Sharpness-aware quantization for deep neural networks.
\newblock \emph{arXiv preprint arXiv:2111.12273}, 2021{\natexlab{b}}.

\bibitem[Liu et~al.(2022{\natexlab{a}})Liu, Mai, Chen, Hsieh, and You]{liu2022towards}
Liu, Y., Mai, S., Chen, X., Hsieh, C.-J., and You, Y.
\newblock Towards efficient and scalable sharpness-aware minimization.
\newblock In \emph{IEEE/CVF Conference on Computer Vision and Pattern Recognition}, 2022{\natexlab{a}}.

\bibitem[Liu et~al.(2022{\natexlab{b}})Liu, Mai, Cheng, Chen, Hsieh, and You]{liu2022random}
Liu, Y., Mai, S., Cheng, M., Chen, X., Hsieh, C.-J., and You, Y.
\newblock Random sharpness-aware minimization.
\newblock In \emph{Advances in Neural Information Processing Systems}, 2022{\natexlab{b}}.

\bibitem[Lopez-Paz \& Ranzato(2017)Lopez-Paz and Ranzato]{lopez2017gradient}
Lopez-Paz, D. and Ranzato, M.
\newblock Gradient episodic memory for continual learning.
\newblock In \emph{Advances in Neural Information Processing Systems}, 2017.

\bibitem[Mehta et~al.(2023)Mehta, Patil, Chandar, and Strubell]{mehta2021empirical}
Mehta, S.~V., Patil, D., Chandar, S., and Strubell, E.
\newblock An empirical investigation of the role of pre-training in lifelong learning.
\newblock \emph{Journal of Machine Learning Research}, 2023.

\bibitem[Mordido et~al.(2022)Mordido, Chandar, and Leduc-Primeau]{mordido2022sharpness}
Mordido, G., Chandar, S., and Leduc-Primeau, F.
\newblock Sharpness-aware training for accurate inference on noisy {DNN} accelerators.
\newblock \emph{arXiv preprint arXiv:2211.11561}, 2022.

\bibitem[Neyshabur et~al.(2017)Neyshabur, Bhojanapalli, McAllester, and Srebro]{neyshabur2017exploring}
Neyshabur, B., Bhojanapalli, S., McAllester, D., and Srebro, N.
\newblock Exploring generalization in deep learning.
\newblock \emph{Advances in Neural Information Processing Systems}, 2017.

\bibitem[Spoon et~al.(2021)Spoon, Tsai, Chen, Rasch, Ambrogio, Mackin, Fasoli, Friz, Narayanan, Stanisavljevic, and Burr]{10.3389/fncom.2021.675741}
Spoon, K., Tsai, H., Chen, A., Rasch, M.~J., Ambrogio, S., Mackin, C., Fasoli, A., Friz, A.~M., Narayanan, P., Stanisavljevic, M., and Burr, G.~W.
\newblock Toward software-equivalent accuracy on {Transformer}-based deep neural networks with analog memory devices.
\newblock \emph{Frontiers in Computational Neuroscience}, 2021.

\bibitem[Stutz et~al.(2021)Stutz, Hein, and Schiele]{Stutz_2021_ICCV}
Stutz, D., Hein, M., and Schiele, B.
\newblock Relating adversarially robust generalization to flat minima.
\newblock In \emph{International Conference on Computer Vision}, 2021.

\bibitem[Vaswani et~al.(2017)Vaswani, Shazeer, Parmar, Uszkoreit, Jones, Gomez, Kaiser, and Polosukhin]{vaswani2017attention}
Vaswani, A., Shazeer, N., Parmar, N., Uszkoreit, J., Jones, L., Gomez, A.~N., Kaiser, {\L}., and Polosukhin, I.
\newblock Attention is all you need.
\newblock \emph{Advances in Neural Information Processing Systems}, 2017.

\bibitem[Xu et~al.(2013)Xu, Niu, Muralimanohar, Jouppi, and Xie]{10.1145/2463209.2488867}
Xu, C., Niu, D., Muralimanohar, N., Jouppi, N.~P., and Xie, Y.
\newblock Understanding the trade-offs in multi-level cell {ReRAM} memory design.
\newblock In \emph{Annual Design Automation Conference}, 2013.

\bibitem[Zagoruyko \& Komodakis(2016)Zagoruyko and Komodakis]{zagoruyko2016wide}
Zagoruyko, S. and Komodakis, N.
\newblock Wide residual networks.
\newblock \emph{arXiv preprint arXiv:1605.07146}, 2016.

\bibitem[Zhang et~al.(2019)Zhang, Lucas, Ba, and Hinton]{zhang2019lookahead}
Zhang, M., Lucas, J., Ba, J., and Hinton, G.~E.
\newblock Lookahead optimizer: k steps forward, 1 step back.
\newblock \emph{Advances in Neural Information Processing Systems}, 2019.

\bibitem[Zhou et~al.(2022)Zhou, Liu, Zhang, and Chen]{zhou2021delta}
Zhou, W., Liu, F., Zhang, H., and Chen, M.
\newblock Sharpness-aware minimization with dynamic reweighting.
\newblock In \emph{Findings of the Association for Computational Linguistics: EMNLP}, 2022.

\bibitem[Zhuang et~al.(2022)Zhuang, Gong, Yuan, Cui, Adam, Dvornek, sekhar tatikonda, s~Duncan, and Liu]{zhuang2022surrogate}
Zhuang, J., Gong, B., Yuan, L., Cui, Y., Adam, H., Dvornek, N.~C., sekhar tatikonda, s~Duncan, J., and Liu, T.
\newblock Surrogate gap minimization improves sharpness-aware training.
\newblock In \emph{International Conference on Learning Representations}, 2022.

\end{thebibliography}
\bibliographystyle{icml2024}

\newpage
\appendix
\onecolumn

\section{Appendix}

Here, we provide additional discussions (Section \ref{sec:additional_discussions}) and more information on the Lookahead-SAM baseline (Section \ref{sec:lookahead_sam}). Moreover, we present further details on the training procedures (Sections \ref{sec:app_training_details} and \ref{sec:hyperparameter_tuning}) and the lifelong learning setup (Section \ref{sec:app_lll_exp_details}). We also provide additional sensitivity analysis across all tested models (Section \ref{sec:app_sensitivity}). Lastly, we present further comparisons on additional training setups (Section \ref{sec:app_asam_setup}) and variants (Section \ref{sec:m_sam}), and provide a speed of convergence analysis (Section \ref{sec:speed_convergence}).

\subsection{Additional discussions}\label{sec:additional_discussions}

In this section, we further discuss the limitations of our work (Section \ref{sec:limitations}) as well as additional studies to better understand the behavior of Lookbehind. In particular, we showcase the advantage of going farther away from the original solution as performing multiple ascent steps instead of staying within a neighborhood size $\rho$ (Section \ref{sec:rho_over_k}), and how the values of $\alpha^*$ evolve during training (Section \ref{sec:adaptive_alpha_values}).

\subsubsection{Limitations}\label{sec:limitations}

One drawback of our approach is the introduction of two new hyperparameters to SAM/ASAM. This was partially addressed in Section \ref{sec:adaptive_alpha} by removing the need to fine-tune $\alpha$. 
Nevertheless, even with the adaptive $\alpha$ variant, our method still introduces one more hyperparameter. Since tuning hyperparameters requires more compute, the comparison with baselines with less hyperparameters is only reasonable to the extent that the baselines are not subject to computational constraints that might limit their performance, \textit{e.g.} by not training for long enough. 
However, we argue that this was not the case in our experimental setup, and additional training would be unlikely to improve the performances reported for the SAM/ASAM baselines. To corroborate this, we show the average number of epochs at which the best SAM and ASAM baseline configurations achieved the best validation accuracy in Table \ref{tab:epochs_sam_asam}. We observe that the best-performing model checkpoints were not completed at the very end of training (\textit{e.g.} last epoch) across our experimental setup, suggesting there was prior performance saturation before training finished. 

\begin{table*}[h]
\caption{Average number of epochs at which the SAM and ASAM baselines achieved the best validation accuracy across the different models and datasets. The models were trained for a total of 200 epochs for CIFAR-10/100 and 90 epochs for ImageNet.}
\begin{center}
\resizebox{0.8\textwidth}{!}{
\begin{tabular}{|l|c|c||c|c||c|}
\hline
Dataset & \multicolumn{2}{c||}{CIFAR-10} & \multicolumn{2}{c||}{CIFAR-100} & ImageNet\\
Model & ResNet-34 & WRN-28-2 & ResNet-50 & WRN-28-10 & ResNet-18\\
\hline\hline
SAM & $164.66_{\pm 9.87}$  & $141.33_{\pm 15.45}$  & $167.66_{\pm 21.06}$ & $172.00_{\pm 9.00}$ & $83.66_{\pm 2.05}$\\
\hline
ASAM & $164.00_{\pm 12.83}$ & $158.66_{\pm 29.45}$ & $178.66_{\pm 3.77}$ & $179.50_{\pm 4.50}$ & $86.00 _{\pm 2.16}$\\
\hline
\end{tabular}}
\end{center}
\label{tab:epochs_sam_asam}
\end{table*}

\subsubsection{Staying within a neighborhood size $\rho$ or $\rho/k$}
\label{sec:rho_over_k}

\begin{figure}[t]
    \centering
    \begin{subfigure}[b]{1.\textwidth}
        \centering
        \includegraphics[width=0.7\linewidth]{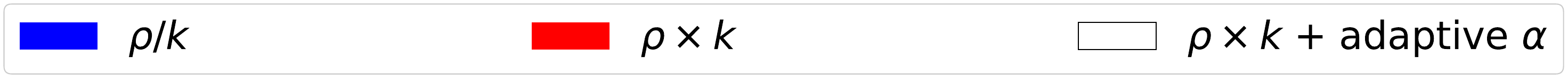}
    \end{subfigure}
    \begin{subfigure}[b]{0.5\textwidth}
        \centering
        \includegraphics[width=0.475\linewidth]{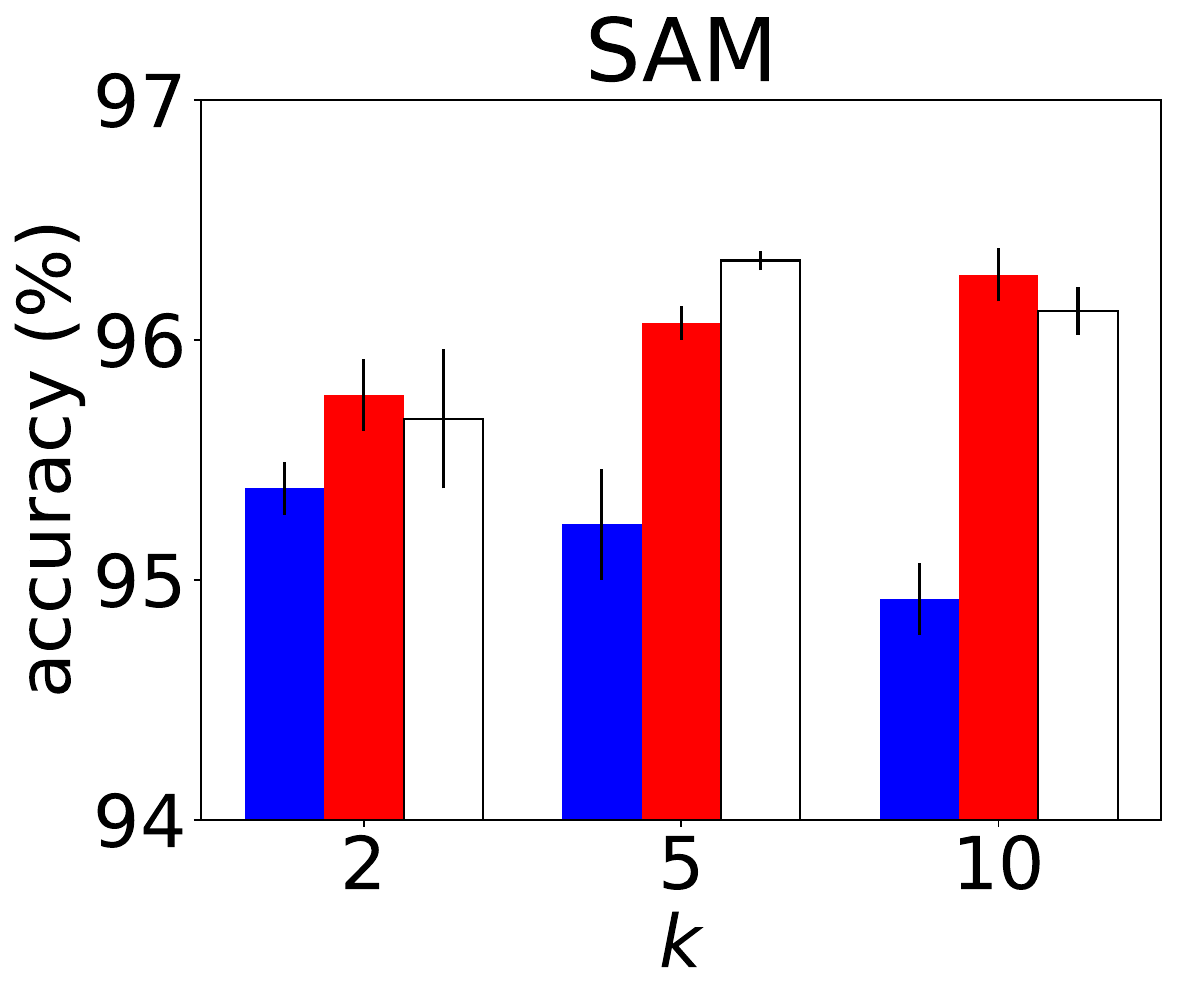}
        \hfill
        \includegraphics[width=0.475\linewidth]{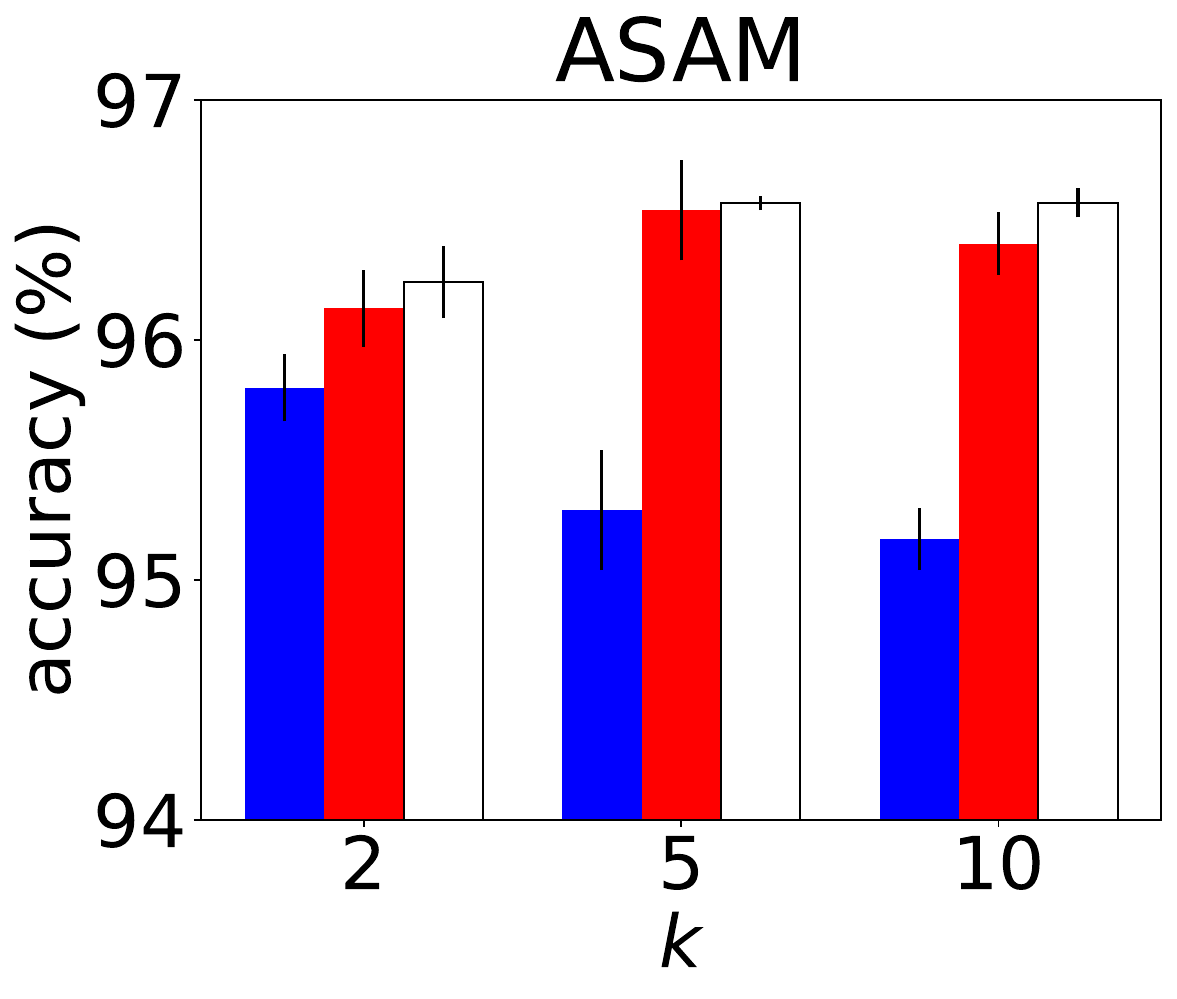}
    \end{subfigure}
    \caption{Comparison of generalization performance (validation accuracy \%) on ResNet-34 trained on CIFAR-10 between staying up to a neighborhood $\rho$ or $\rho \times k$. We also plot the performance of adaptive $\alpha$ in the latter setting.
    }
    \label{fig:ablation_rho_over_k}
\end{figure}

As a wrapper to SAM methods, Lookbehind's practicality is enhanced when there is no need to re-tune the default $\rho$ of the sharpness-aware minimizer. To study this, we used the default $\rho$ suggested by SAM and ASAM and investigated if staying within a neighborhood $\rho$ of the original solution is more advantageous than increasing the neighborhood up to $\rho \times k$, as presented so far throughout our paper. For this new variant, we reduce the neighborhood size to $\rho/k$ as the step size for each ascent step. Hence, after $k$ ascent steps we will be at a maximum distance $\rho$ from the original point if all gradients align. We also remove linear interpolation and simply set the descent step size to $\eta$. Results using the default $\rho$ for SAM and ASAM are presented in Figure \ref{fig:ablation_rho_over_k}.

We observe that going farther away as we perform the ascent steps consistently outperforms staying within a neighborhood $\rho$ of the original solution. In other words, $\rho \times k$ is better than $\rho/k$ when using the default $\rho$ of SAM and ASAM. This is a convenient insight since we show that tuning the hyperparameter $\rho$ is not necessary when using the former setting. Moreover, this also allows us to learn $\alpha$ dynamically, which is shown to enhance performances in some settings. This suggests that it is beneficial to not only "look behind" within a neighborhood of $\rho \times k$, but also that taking a dynamic descent step size to perform the final update based on the alignment of the aggregated gradients is an effective way of enhancing performance across different $k$.

\subsubsection{Change of $\alpha^*$ during training}
\label{sec:adaptive_alpha_values}

\begin{figure}
     \centering
          \begin{subfigure}[b]{0.49\textwidth}
         \centering
         \includegraphics[width=1.\textwidth]{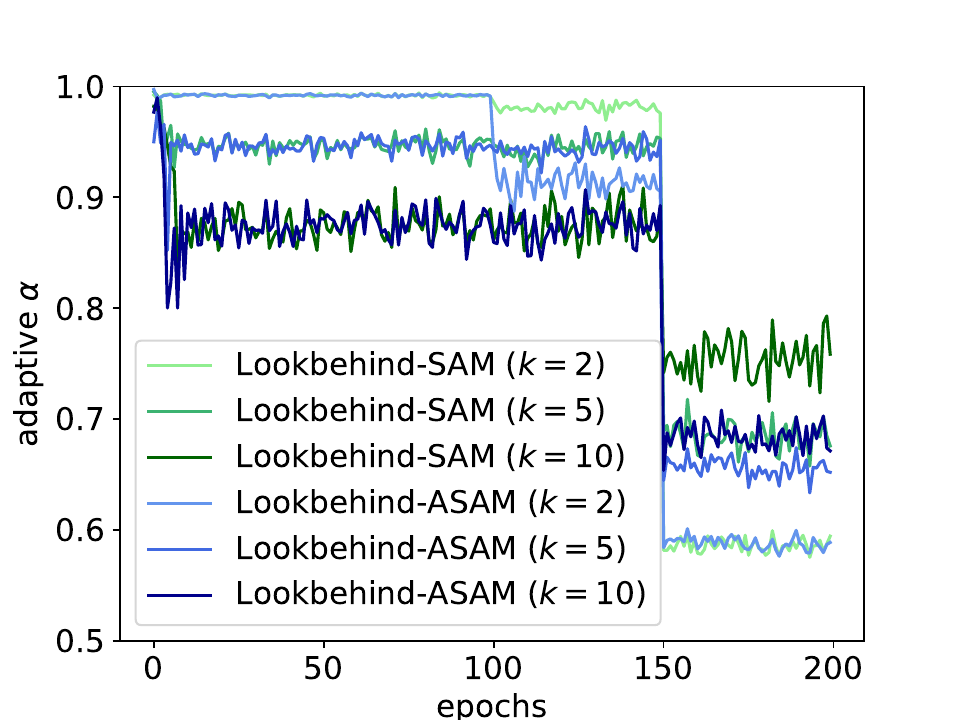}
         \caption{ResNet-34 on CIFAR-10.}
     \end{subfigure}
     \hfill
     \begin{subfigure}[b]{0.49\textwidth}
         \centering
         \includegraphics[width=1.\textwidth]{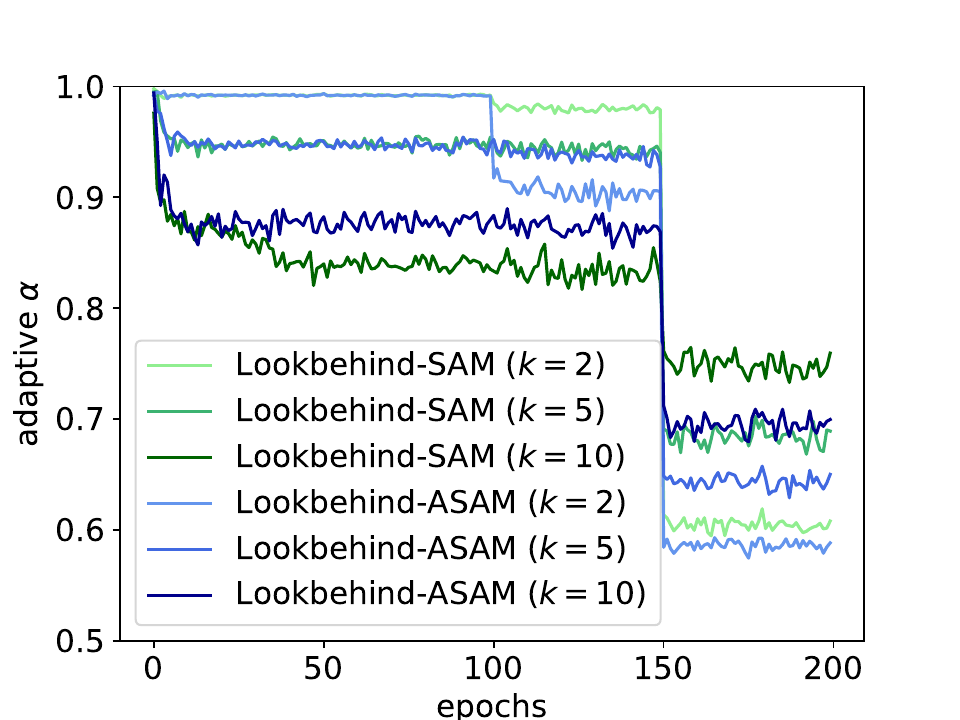}
         \caption{ResNet-50 on CIFAR-100.}
     \end{subfigure}
     \hfill
     \begin{subfigure}[b]{0.48\textwidth}
         \centering
         \includegraphics[width=1.\textwidth]{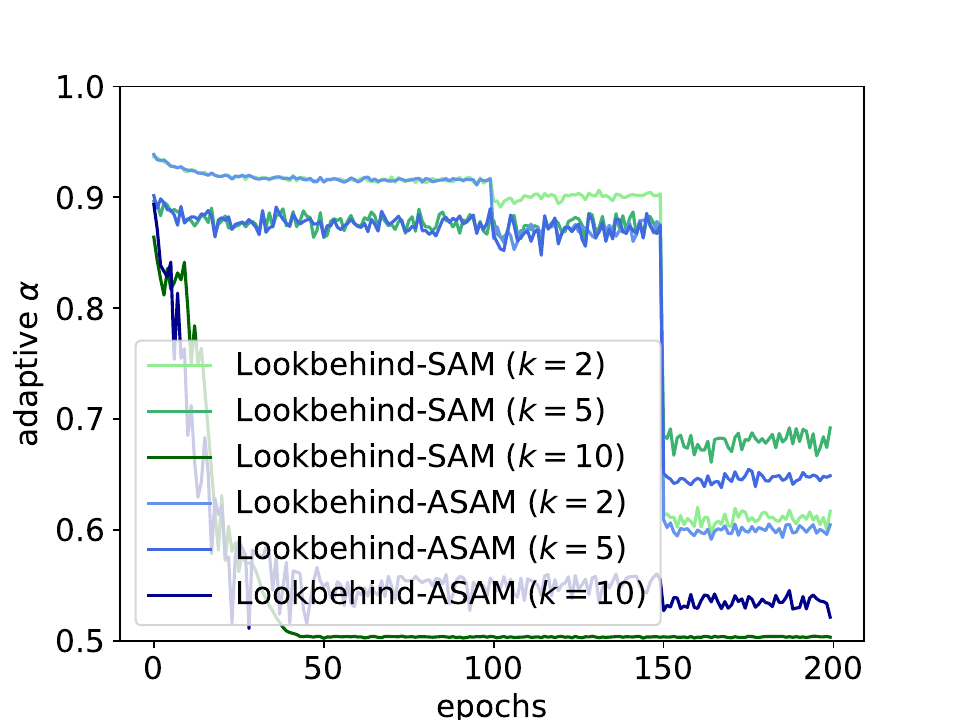}
         \caption{WRN-28-2 on CIFAR-10.}
     \end{subfigure}
     \hfill
     \begin{subfigure}[b]{0.48\textwidth}
         \centering
         \includegraphics[width=1.\textwidth]{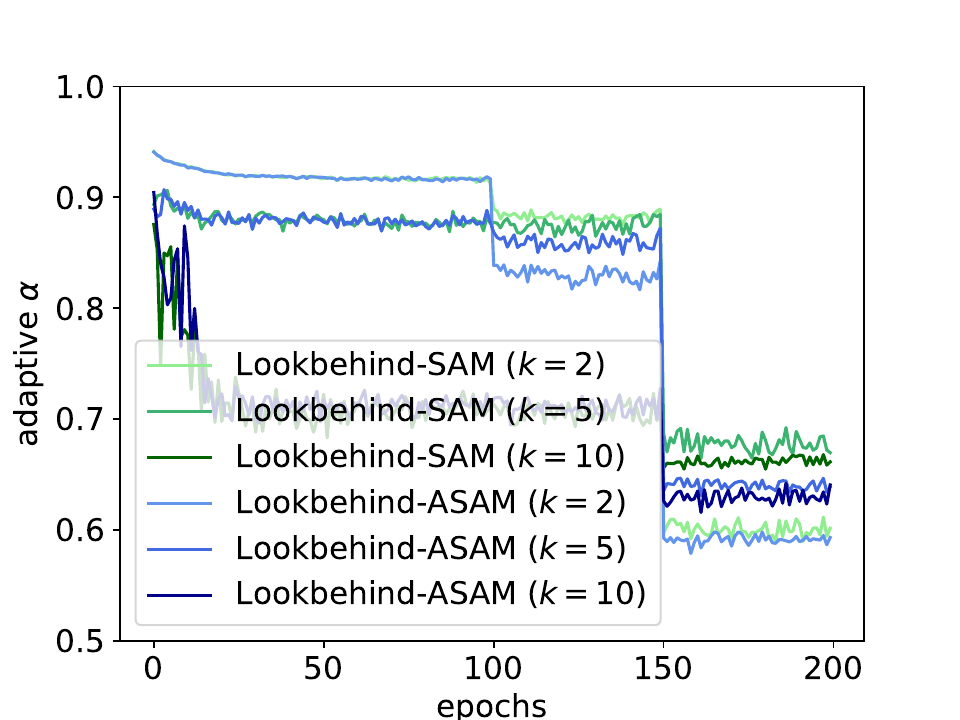}
         \caption{WRN-28-10 on CIFAR-100.}
     \end{subfigure}
     \hfill
        \caption{Analysis of how adaptive $\alpha$ evolves during training.}
        \label{fig:adaptive_alpha_evolution}
\end{figure}

We show how adaptive $\alpha$ changes throughout training in Figure \ref{fig:adaptive_alpha_evolution}. We notice an expected trend based on the values of $k$, with higher $k$ leading to lower $\alpha^*$ due to less gradient alignment. Even though $\alpha$ is independent of the inner step learning rate $\eta$, we are decreasing $\eta$ by a factor of 10 every 50 epochs in our training setup, which leads to drastic changes in model performance and loss landscape. This in turn seems to lead to an increase in the misalignment of the aggregated gradients which decreases the adaptive $\alpha$ values later on in training.

\subsection{Lookahead-SAM}
\label{sec:lookahead_sam}

\label{sec:LH+SAM}
\begin{figure}
\centering
\begin{minipage}{.6\textwidth}
\centering
\begin{algorithm}[H]
\caption*{\textbf{Algorithm 2} Lookahead-SAM}
\begin{algorithmic}[1]
\REQUIRE Initial parameters $\phi_0$, loss function $L$, inner steps $k$, slow weights step
size $\alpha$, fast weights step
size $\eta$, neighborhood size $\rho$, training set $D$
\FOR{$t = 1, 2, \ldots$}
\STATE $\phi_{t,0} \gets \phi_{t-1}$
\FOR{$i = 1, 2, \ldots, k$}
\STATE Sample mini-batch $d \sim D$
\STATE $\epsilon \gets \rho \dfrac{\nabla L_d(\phi_{t,i-1})}{\|\nabla L_d(\phi_{t,i-1})\|_2}$
\STATE $\phi_{t,i} \gets \phi_{t,i-1} - \eta \nabla_{L_{d}} (\phi_{t,i-1} + \epsilon)$
\ENDFOR
\STATE $\phi_{t} \gets \phi_{t-1} + \alpha(\phi_{t,k} - \phi_{t-1})$
\ENDFOR
\STATE {\bfseries return} $\phi$
\end{algorithmic}
\end{algorithm}
\vfill
\end{minipage}
\begin{minipage}{.39\textwidth}
    \begin{subfigure}[b]{1.\textwidth}
        \includegraphics[width=1.\linewidth]{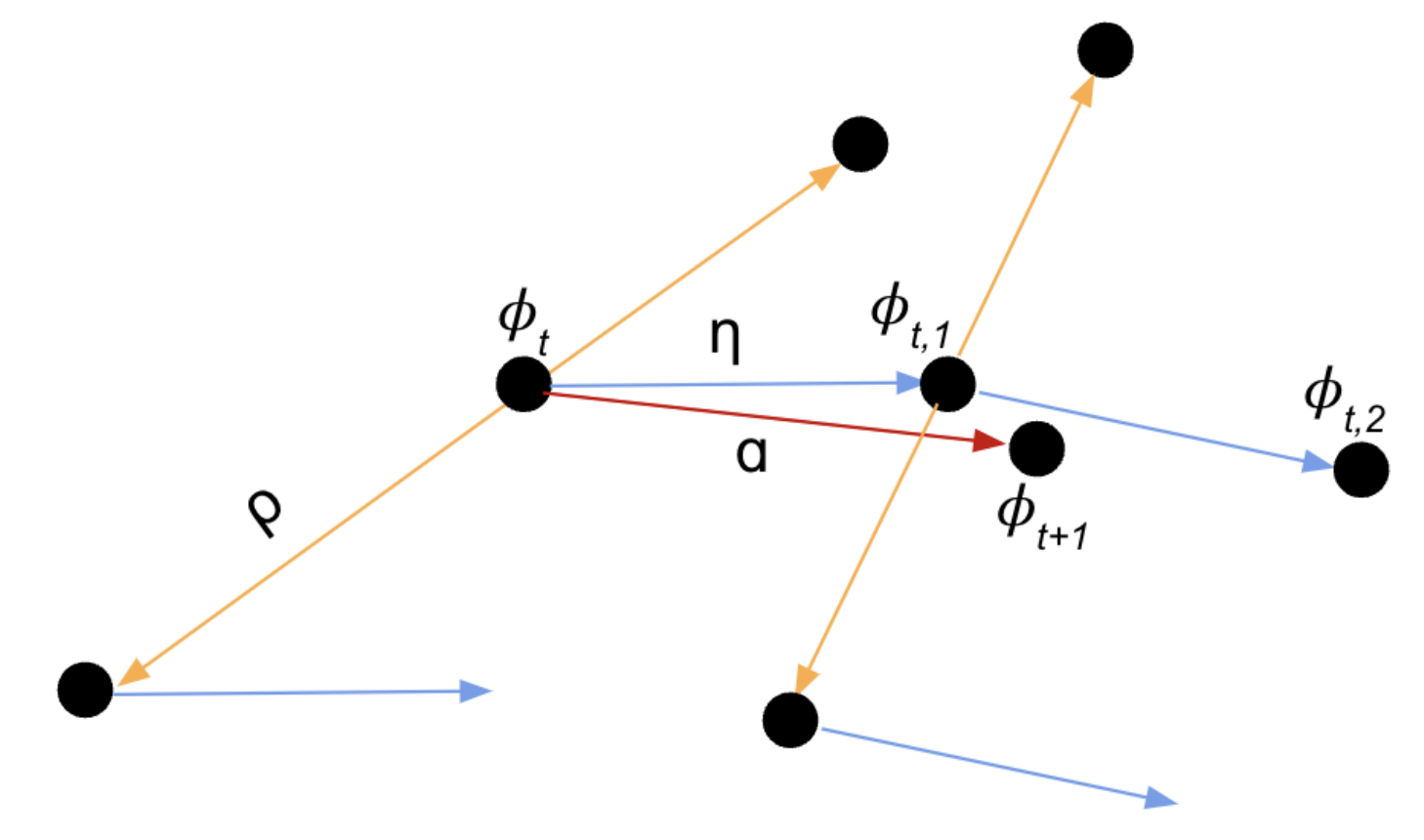}
        \label{fig:lookahead}
    \end{subfigure}
\end{minipage}
\caption{Combination of Lookahead with SAM.}\label{fig:algs}
\end{figure}

Lookahead \citep{zhang2019lookahead} was introduced to reduce variance during training, with the end goal of improving performance and robustness to hyper-parameter settings.
Given an optimizer, Lookahead uses slow and fast weights to improve its training stability. The algorithm "looks ahead" by updating the fast weights $k$ times in an inner loop, while the slow weights are updated by performing a linear interpolation to the final fast weights (after the inner loop ends).  In our analysis and experiments, we use Lookahead with sharpness-aware methods by applying single-step SAM and ASAM as the inner optimizers. The main goal of these baselines is to use Lookahead to stabilize sharpness-aware optimizers when training with large $\rho$. An illustration of Lookahead-SAM is presented in Figure \ref{fig:algs} (right). 

Similarly to our method, Lookahead-SAM uses slow weights ($\phi_{t}$, $\phi_{t+1}, \cdots$) and fast weights ($\phi_{t, 1}$, $\cdots$, $\phi_{t, k}$). However, the slow weights are updated after each SAM update (composed of a single ascent and descent step), while the slow weights are updated toward the fast weights through linear interpolation after $k$ steps ($\phi_{t+1}$). In contrast, Lookbehind-SAM's fast and slow weights are obtained during a given iteration. In particular, while the fast weights are updated as we "look behind", the slow weights are updated after $k$ ascent steps are performed (c.f. Figure \ref{fig:sams}).

The pseudo-code for combining Lookahead with SAM is presented in Figure \ref{fig:algs} (left). Just like Lookahead, Lookahead-SAM maintains a set of slow weights and fast weights, which are synchronized at the beginning of every outer step (line 2). Then, the fast weights are updated $k$ times (looking forward) using a standard SAM update with a single ascent (line 5) and descent step (line 6). After $k$ such SAM steps, the slow weights are updated by linearly interpolating to the final fast weights (line 8) (1 step back). It is worth noting that a new minibatch is sampled at every inner step (line 4). 
Combining Lookahead with ASAM follows the same procedure, except using the component-wise rescaling \eqref{eq:ASAM} in line 5. 

Although Lookbehind-SAM and Lookahead-SAM share a similar nature, they exhibit notable distinctions. Firstly, in addition to synchronizing the fast weights, Lookbehind also synchronizes the perturbed fast weights. Furthermore, the minibatch is sampled before the inner loop. Moreover, at each inner step, Lookbehind performs $k$ ascent steps of SAM.
The distinction between the two algorithms leads to divergent behavior in the training objective and is related to Lookahead-SAM and Lookbehind-SAM having different goals: while Lookahead-SAM aims at stabilizing single-step SAM with large neighborhood sizes $\rho$, Lookbehind aims to perform multiple ascent steps while maintaining a good balance between sharpness and training accuracy. 

In other words, Lookbehind focuses on curbing the variance arising from gradients gathered during multiple ascent steps within a single iteration. In contrast, Lookahead-SAM targets variance stemming from sequential descent steps performed across iterations. Hence, our goal is to reduce the variance of looking behind, not ahead. 

\subsection{Training details}\label{sec:app_training_details}

For CIFAR-10/100, we trained each model for 200 epochs with a batch size of 128, starting with a learning rate of 0.1 and dividing it by 10 every 50 epochs. For ResNet-18 trained from scratch on ImageNet, we used 1000 classes and an image size of 224x224 and trained each model for 90 epochs with a batch size of 400, starting with a learning rate of 0.1 and dividing it by 10 every 30 epochs. All models were trained using SGD with momentum set to 0.9 and weight decay of 1e-4. For the fine-tuning experiments on ViT-B, we trained for 15 epochs using a batch size of 450 to maximize resource utilization and a learning rate of 0.001. We trained the CIFAR-10/100 models using one RTX8000 NVIDIA GPU and 1 CPU core, and the ImageNet models using one A100 GPU (with 40 and 80 GB of memory for training from scratch and fine-tuning, respectively) and 6 CPU cores. For the machine translation experiments, we used the setup in \url{https://github.com/facebookresearch/fairseq/} and trained the models using one RTX8000 NVIDIA GPU with 6 CPU cores.

\subsection{Hyperparameter search}\label{sec:hyperparameter_tuning}

For Table \ref{tab:generalization}, we perform hyperparameter search for $\rho \in \{0.005, 0.01, 0.02, 0.05, 0.1, 0.2, 0.5, 1.0, 2.0\}$ for the vanilla SAM and ASAM baselines trained on CIFAR-10/100, and report the validation results with the best $\rho$. For the rest of the methods, we used the default $\rho$, \textit{i.e.} as presented in the original SAM \cite{foret2021sharpnessaware} and ASAM \cite{kwon2021asam} papers. Particularly, we used $\rho$ of 0.05, 0.1, and 0.05 for SAM and 0.5, 1.0, and 1.0 for ASAM when training on CIFAR-10, CIFAR-100, and ImageNet, respectively. For CIFAR-10/100, we use $k \in \{2, 5, 10\}$ and $\alpha \in \{0.2, 0.5, 0.8\}$ (when applicable) for the multiple step methods. For ImageNet, we use $k = 2$ and $\alpha \in \{0.2, 0.5, 0.8\}$ (when applicable).

For Figure \ref{fig:sharpness_rho}, we report the best $k$ and $\alpha$ configurations for all methods, \textit{i.e.} with the lowest sharpness at the highest $r$.

For Figure \ref{fig:noisy_weights}, we report the most robust model using $k \in \{2, 5, 10\}$ and $\alpha \in \{0.2, 0.5, 0.8\}$ for CIFAR-10/100. For ImageNet, we use $k = 2$ and $\alpha \in \{0.2, 0.5, 0.8\}$. For the SAM and ASAM baselines, we pick the most robust $\rho \in \{0.05, 0.1, 0.2, 0.5\}$ and $\rho \in \{0.5, 1.0, 2.0, 5.0\}$, respectively.

For Figure \ref{fig:ablation} we report the default neighborhood sizes for the SAM ($\rho = 0.05$ and $0.1$ for CIFAR-10 and CIFAR-100, respectively) and ASAM baselines ($\rho = 0.5$ and $1.0$ for CIFAR-10 and CIFAR-100, respectively). We show the best hyper-parameter configuration over $k \in \{2, 5, 10\}$ and $\alpha \in \{0.2, 0.5, 0.8\}$ for Lookbehind and Lookahead, and $k \in \{2, 5, 10\}$ for Multistep.

For Figure \ref{fig:rho_test_accuracies}, we report the best $\alpha$ configuration for Lookahead and Lookbehind.

For Table \ref{tab:transformers}'s image classification experiments, we report and use the default $\rho$ for SAM across all the models. For ASAM, we report the best $\rho$ for each method over $\rho \in \{0.1, 0.2, 0.5\}$ since the default $\rho$ did not outperform the pre-trained model performance. All reported ASAM models achieved the best performance with $\rho=0.1$.

For Table \ref{tab:transformers}'s machine translation experiments, we report the best $\rho$ using the same search space as the one used by \citet{kwon2021asam} for a fair comparison, \textit{i.e.} $\rho \in \{0.005, 0.01, 0.02, \ldots, 0.5, 1.0, 2.0\}$. We report the results for the Adam, SAM, and ASAM baselines as presented in their paper.

For both experiments reported in Table \ref{tab:transformers} we do not do any hyperparameter search over $\alpha$, simply setting it to 0.8 since it was shown to work well with $k=2$ in our previous experiments (c.f. Figure \ref{fig:lookbehind_app}).

\subsection{Lifelong learning}\label{sec:app_lll_exp_details}

We replicated the experimental setup from Lookahead-MAML \citep{gupta2020look} and report the results for all baselines where the models were trained for $10$ epochs per task. Additionally, we combined the different methods with episodic replay (ER) \citep{chaudhry2019tiny}, which maintains a memory of a subset of the data from each task and uses it as a replay buffer while training on new tasks. We test both settings (with and without ER) in our experiments. We used two datasets: Split-CIFAR100 and Split-TinyImageNet. The Split-CIFAR100 benchmark is designed by splitting the 100 classes in CIFAR-100 into 20 5-way classification tasks. Similarly, Split-TinyImageNet is designed by splitting 200 classes into 40 5-way classification tasks. In both cases, the task identities are provided to the model along with the dataset. Each model has multi-head outputs, \textit{i.e.} each task has a separate classifier.

We provide the grid search details for finding the best set of hyper-parameters for both datasets and all baselines in Table \ref{tab:gridsearch_lll}. We train the model on the training set and report the best hyper-parameters based on the highest accuracy on the test set in Table \ref{tab:best_hyp_lll}. Here, we report the hyper-parameter set for each method (with or without ER) as follows: 

\begin{itemize}
    \item SGD: $\{\eta\}$
    \item SAM: $\{\eta, \rho\}$
    \item Multistep-SAM: $\{\eta, \rho, k\}$
    \item Lookbehind-SAM: $\{\eta, \rho, k, \alpha\}$
    \item Lookbehind-C-MAML: $\{\eta, \rho, k, \alpha\}$
\end{itemize}

We refer to \citet{gupta2020look} for the best hyper-parameters of Lookahead-C-MAML. We evaluated all models using the following metrics: 
    \begin{itemize}
        \item \textbf{Average accuracy} \citep{lopez2017gradient}: the average performance of the model across all the previous tasks is defined by
        $
        \frac{1}{t}\sum\limits_{\tau=1}^t a_{t,\tau}
        $, where $a_{t,\tau}$ is the accuracy on the test set of task $\tau$ when the current task is $t$. 
        \item \textbf{Forgetting} \citep{chaudhry2018riemannian}:  the average forgetting that occurs after the model is trained on several tasks is computed by
        $
        \frac{1}{t-1}\sum\limits_{\tau=1}^{t-1} \max_{t' \in \{1,...,t-1\}} (a_{t',\tau} - a_{t,\tau})
        $, where $t$ represents the latest task. 
    \end{itemize}

We report the average accuracy and forgetting after the models were trained on all tasks for both datasets.

\begin{table*}[h]\caption{Details on the hyper-parameter grid search used for the lifelong learning experiments.}\label{tab:gridsearch_lll}
\centering
\begin{tabular}{@{}cc@{}}

\toprule
Hyper-parameters           & Values \\ \midrule
step size ($\eta$)            & $\{0.3, 0.1, 0.03, 0.01, 0.003, 0.001, 0.0003, 0.0001, 0.00003, 0.00001\}$                 \\
inner steps ($k$)           & $\{2, 5, 10\}$                    \\
outer step size ($\alpha$)           & $\{0.1, 0.2, 0.5, 0.8, 1.0\}$                       \\
neighborhood size ($\rho$)           & $\{0.005, 0.01, 0.05, 0.1\}$                       \\
\bottomrule
\end{tabular}
\end{table*}

\begin{table*}[h]\caption{Best hyper-parameter settings for the different lifelong learning methods.}\label{tab:best_hyp_lll}
\centering
\begin{tabular}{@{}ccc@{}}
\toprule
Methods    & Split-CIFAR100 & Split-TinyImagenet \\ \midrule
SGD                 & $\{0.03\}$                & $\{0.03\}$                    \\
SAM                 & $\{0.03, 0.05\}$          & $\{0.03, 0.05\}$              \\
Multistep-SAM       & $\{0.01, 0.01, 2\}$       & $\{0.03, 0.05, 2\}$           \\
Lookbehind-SAM      &     $\{0.1, 0.05, 10, 0.1\}$                    & $\{0.01, 0.05, 10, 0.1\}$     \\
ER + SAM            &          $\{0.1, 0.05\}$               & $\{0.03, 0.1\}$               \\
ER + Multistep-SAM  & $\{0.1, 0.05, 10\}$       & $\{0.03, 0.1, 10\}$           \\
ER + Lookbehind-SAM &          $\{0.03, 0.05, 10, 0.2\}$               & $\{0.01, 0.1, 5, 0.5\}$       \\
Lookbehind-C-MAML   & $\{0.03, 0.005, 2, 1\}$   & $\{0.03, 0.1, 2, 1\}$         \\ \bottomrule
\end{tabular}
\end{table*}

The pseudo-code for Lookahead-C-MAML and Lookbehind-C-MAML is presented in Figure \ref{fig:lll_algs}.

\begin{figure}
\centering
\begin{minipage}{.6\textwidth}
\centering
\begin{algorithm}[H]
\caption*{\textbf{Algorithm 3} Lookahead-C-MAML \citep{gupta2020look}}
\begin{algorithmic}[1]
\REQUIRE Initial parameters $\phi^0_0$, inner loss function $\ell$, meta loss function $L$, step
size $\eta$, training set $D_t$ of task $t$, number of epochs $E$
\STATE $j \gets 0$
\STATE $R \gets$ \{\}
\FOR{$t = 1, 2, \ldots$}
\STATE Sample batch $d_t \sim D_t$
\FOR{$e = 1, 2, \ldots, E$}
\FOR{mini-batch $b\textbf{ in }d_t$}
\STATE $k \gets \text{sizeof}(b)$
\STATE $b_m \gets \text{Sample}(R) \cup b$
\FOR{$k' = 0\textbf{ to } k-1$}
\STATE Push $b[k']$ to R
\STATE $\phi^j_{k'+1} \gets \phi^j_{k'} - \eta \nabla_{\phi^j_{k'}} \ell_t(\phi^j_{k'}, b[k'])$
\ENDFOR
\STATE $\phi^{j+1}_0 \gets \phi^{j}_0 - \eta \nabla_{\phi^{j}_0}L_t(\phi^{j}_k, b_m)$
\STATE $j \gets j + 1$
\ENDFOR
\ENDFOR
\ENDFOR
\STATE {\bfseries return} $\phi$
\end{algorithmic}
\end{algorithm}
\vfill
\end{minipage}
\hfill
\begin{minipage}{.95\textwidth}
\centering
\begin{algorithm}[H]
\caption*{\textbf{Algorithm 4} Lookbehind-C-MAML (ours)}
\begin{algorithmic}[1]
\REQUIRE Initial parameters $\phi^0_{0,0}$, inner loss function $\ell$, meta loss function $L$, inner steps $k$, step
size $\eta$, outer step
size $\alpha$, neighborhood size $\rho$, training set $D_t$ of task $t$, number of epochs $E$
\STATE $j \gets 0$
\STATE $R \gets$ \{\}
\FOR{$t = 1, 2, \ldots$}
\STATE $\phi^j_{t,0} \gets \phi^{j}_{t-1,0}$
\STATE Sample batch $d_t \sim D_t$
\FOR{$e = 1, 2, \ldots, E$}
\STATE $\phi'^j_{t,0} \gets \phi^j_{t,0}$
\FOR{$k' = 0\textbf{ to } k-1$}
\STATE Sample mini-batch $b \sim d_t$ of size $k$ without replacement
\STATE $b_m \gets \text{Sample}(R) \cup b$
\STATE Push $b[k']$ to R

\STATE $\epsilon \gets \rho \dfrac{\nabla_{\ell_t}(\phi'^{j}_{t,k'}, b[k'])}{\|\nabla_{\ell_t}(\phi'^{j}_{t,k'}, b[k'])\|_2}$
\STATE $\phi^j_{t,k'+1} \gets \phi^j_{t,k'} - \eta\nabla_{\ell_t}(\phi'^j_{t,k'}+\epsilon, b[k'])$
\ENDFOR
\STATE $\phi^j_{t,k} \gets \phi^{j}_{t,0} + \alpha(\phi^j_{t,k} - \phi^{j}_{t,0})$
\STATE $\epsilon \gets \rho \dfrac{\nabla_{\phi^{j}_{t,0}}L_t(\phi^{j}_{t,k}, b_m)}{\|\nabla_{\phi^{j}_{t,0}}L_t(\phi^{j}_{t,k}, b_m)\|_2}$
\STATE $\phi^{j+1}_{t,0} \gets \phi^{j}_{t,0} - \eta \nabla_{\phi^{j}_{t,0}}L_t(\phi^{j}_{t,0}+\epsilon, b_m)$
\STATE $j \gets j + 1$
\ENDFOR
\ENDFOR
\STATE {\bfseries return}  $\phi$
\end{algorithmic}
\end{algorithm}
\end{minipage}
\caption{Implementations of Lookahead-C-MAML (top) and Lookbehind-C-MAML (bottom).}\label{fig:lll_algs}
\end{figure}

\subsection{Sensitivity to $\alpha$ and $k$}\label{sec:app_sensitivity}

\begin{figure}[h]
     \centering
     \begin{subfigure}[b]{0.2\textwidth}
         \centering
         \includegraphics[width=\textwidth]{Figures/test_accuracies_resnet34.pdf}
         \caption{ResNet-34\\(CIFAR-10)}
     \end{subfigure}
     \hfill
     \begin{subfigure}[b]{0.2\textwidth}
         \centering
         \includegraphics[width=\textwidth]{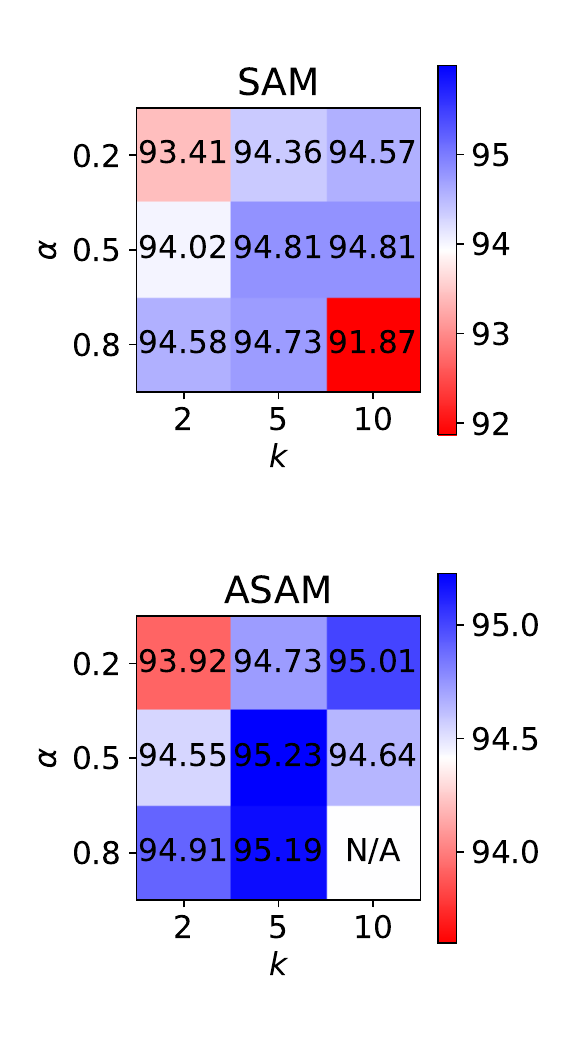}
         \caption{WRN-28-2\\(CIFAR-10)}
     \end{subfigure}
     \hfill
          \begin{subfigure}[b]{0.2\textwidth}
         \centering
         \includegraphics[width=\textwidth]{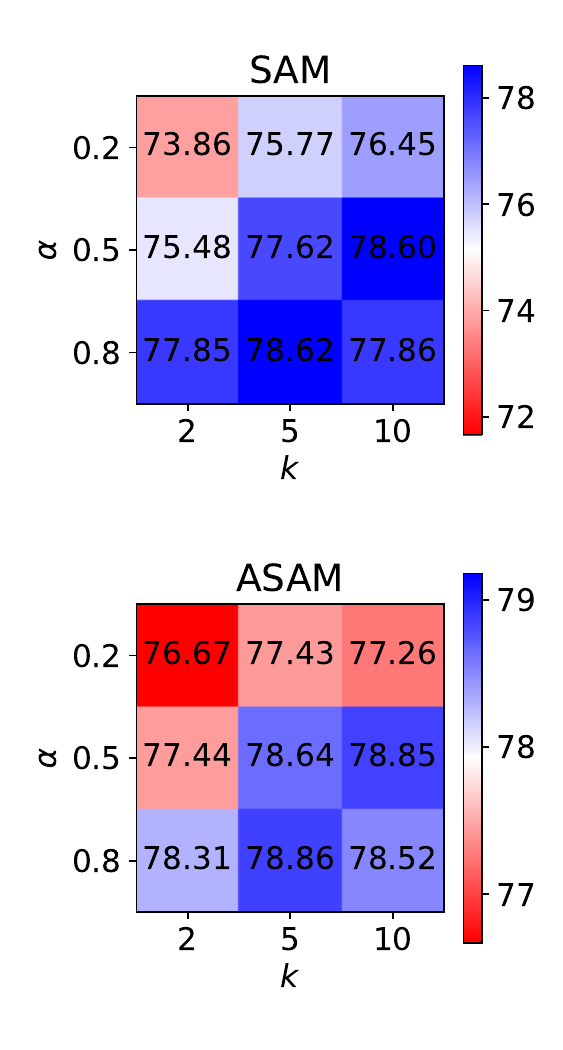}
         \caption{ResNet-50\\(CIFAR-100)}
     \end{subfigure}
     \hfill
     \begin{subfigure}[b]{0.2\textwidth}
         \centering
         \includegraphics[width=\textwidth]{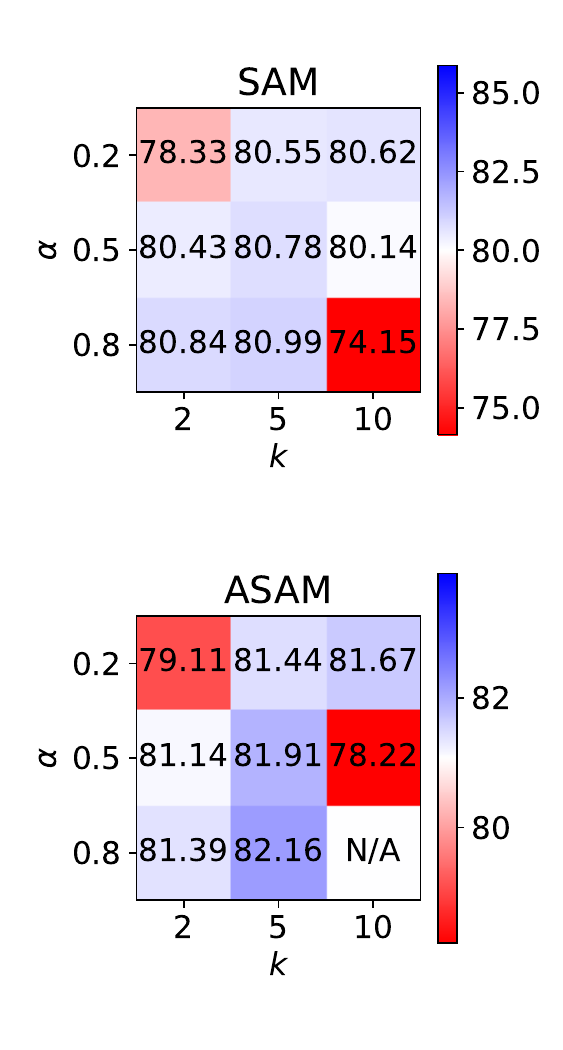}
         \caption{WRN-28-10\\(CIFAR-100)}
     \end{subfigure}
     \hfill
     \begin{subfigure}[b]{0.105\textwidth}
         \centering
         \includegraphics[width=\textwidth]{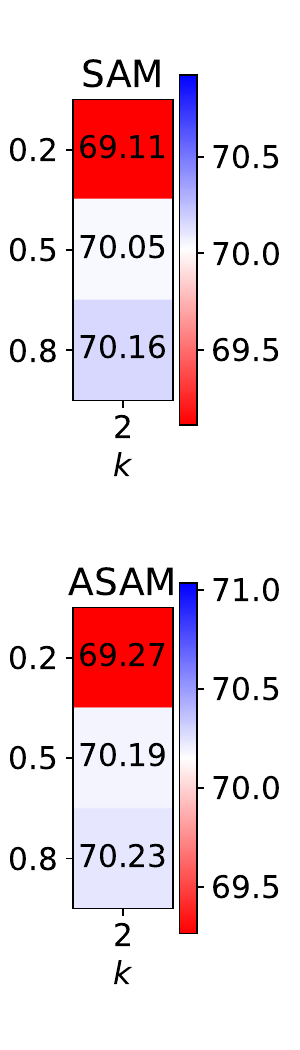}
         \caption{RN-18\\(ImageNet)}
     \end{subfigure}
        \caption{Sensitivity of Lookbehind to $\alpha$ and $k$ when combined with SAM and ASAM in terms of generalization performance (validation accuracy \%). The validation accuracies of the SAM and ASAM variants are presented in the middle of the heatmap (white middle point). All models were trained with the default $\rho$. Blue represents an improvement in terms of validation accuracy over such baselines, while red indicates a degradation in performance. Experiments represented as "N/A" indicate instances where at least one seed failed to converge.}
        \label{fig:lookbehind_app}
\end{figure}

\begin{figure}[h]
     \centering
     \begin{subfigure}[b]{0.2\textwidth}
         \centering
         \includegraphics[width=\textwidth]{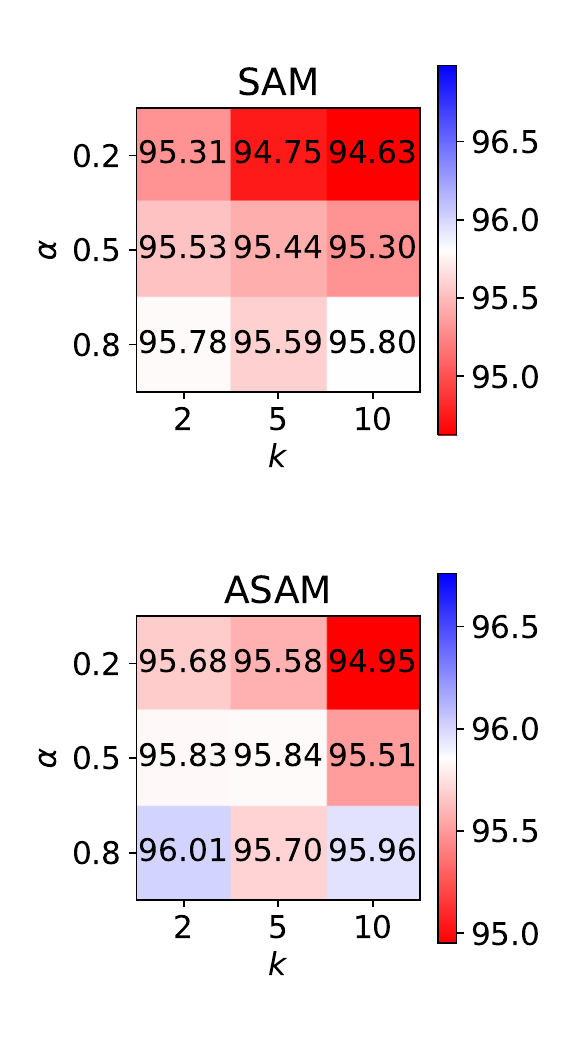}
         \caption{ResNet-34\\(CIFAR-10)}
     \end{subfigure}
     \hfill
     \begin{subfigure}[b]{0.2\textwidth}
         \centering
         \includegraphics[width=\textwidth]{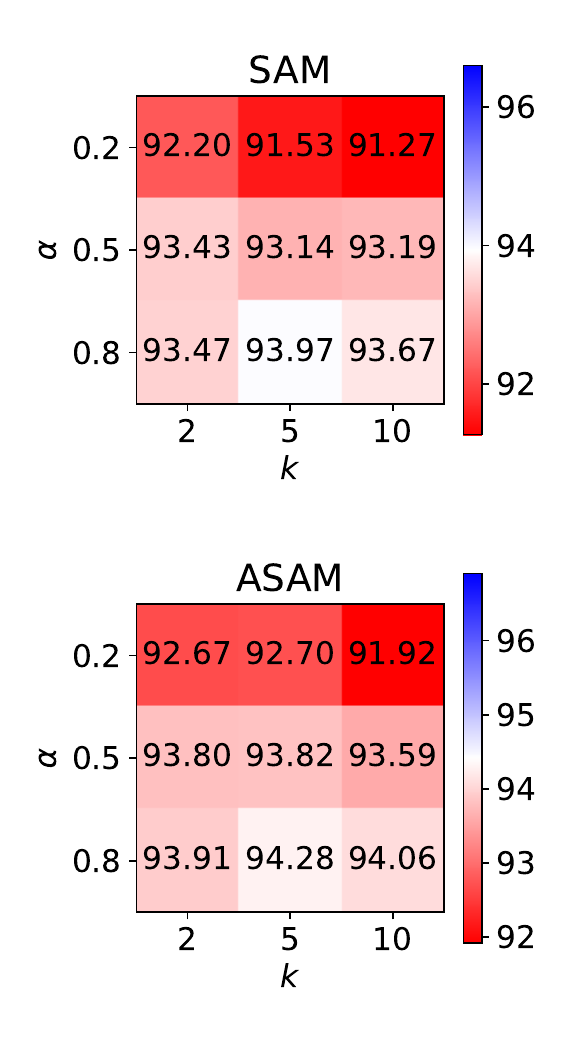}
         \caption{WRN-28-2\\(CIFAR-10)}
     \end{subfigure}
     \hfill
          \begin{subfigure}[b]{0.2\textwidth}
         \centering
         \includegraphics[width=\textwidth]{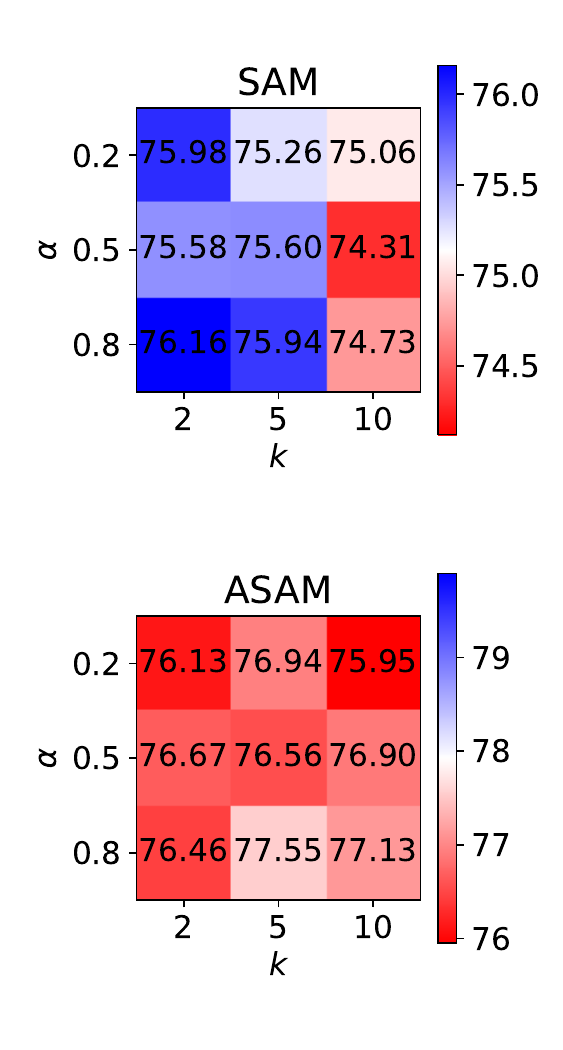}
         \caption{ResNet-50\\(CIFAR-100)}
     \end{subfigure}
     \hfill
     \begin{subfigure}[b]{0.2\textwidth}
         \centering
         \includegraphics[width=\textwidth]{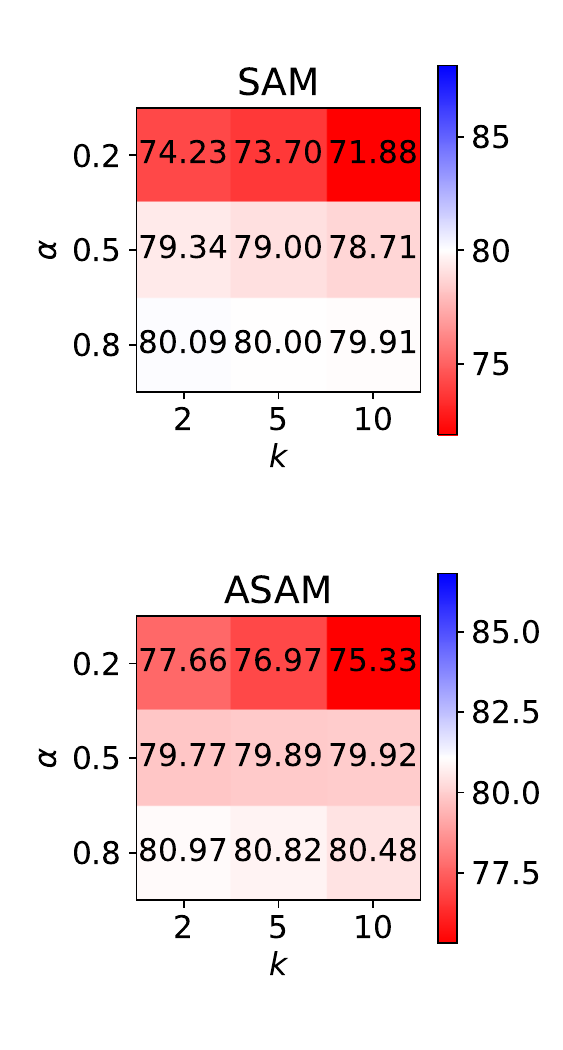}
         \caption{WRN-28-10\\(CIFAR-100)}
     \end{subfigure}
     \hfill
     \begin{subfigure}[b]{0.105\textwidth}
         \centering
         \includegraphics[width=\textwidth]{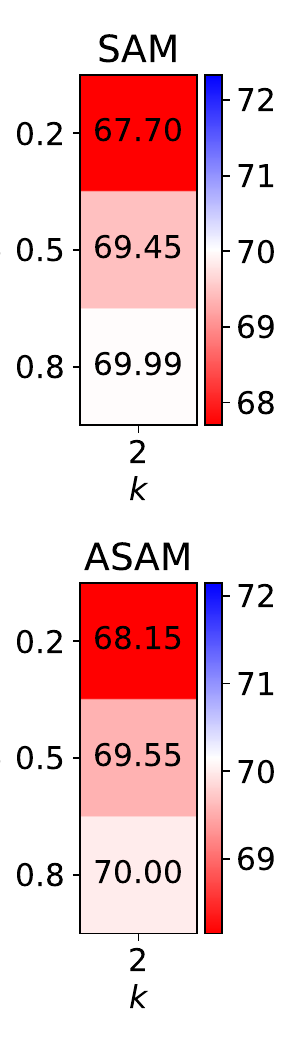}
         \caption{RN-18\\(ImageNet)}
     \end{subfigure}
        \caption{Sensitivity of Lookahead to $\alpha$ and $k$ when combined with SAM and ASAM in terms of generalization performance (validation accuracy \%). The validation accuracies of the SAM and ASAM variants are presented in the middle of the heatmap (white middle point). All models were trained with the default $\rho$. Blue represents an improvement in terms of validation accuracy over such baselines, while red indicates a degradation in performance.}
        \label{fig:lookahead_app}
\end{figure}

We measure the sensitivity to $\alpha$ and $k$ of Lookbehind and Lookahead on additional models in Figures \ref{fig:lookbehind_app} and \ref{fig:lookahead_app}, respectively. Similarly to the sensitivity results presented in the main paper, we observe that Lookbehind is more robust to the choice of $\alpha$ and $k$ and is able to improve on the SAM and ASAM baselines more significantly and consistently than Lookahead.

\subsection{Additional training setups}\label{sec:app_asam_setup}

To further illustrate the superiority of our approach with stronger baselines, we replicated the setup originally used by ASAM described in \cite{kwon2021asam}. The main difference between this new setup and our previous setup is the use of a cosine learning rate scheduler and label smoothing which leads to an increase in generalization performance. To avoid any hyperparameter tuning, we used $k=2$ with an adaptive $\alpha$ for Lookbehind and used SAM and ASAM's default $\rho$ values of 0.1 and 1.0, respectively, as reported in \citet{kwon2021asam}. 

Results over 5 seeds using a WRN-28-2 model trained on CIFAR-100 for 200 epochs are presented in Table \ref{tab:generalization_asam_setup}. We observe that Lookbehind is able to further improve upon the high-accuracy SAM and ASAM baseline models. We note that we recomputed SAM and ASAM's baselines and achieved comparable to the ones reported in \citet{kwon2021asam}: $83.36_{\pm .18}$ and $83.60_{\pm .15}$, respectvely. These results support our conclusions when using the setup used in the experiments in the main paper showcasing the superiority of our method.

\begin{table*}[h]
\caption{Generalization performance (test accuracy \%) of the different methods on WRN-28-10 trained on CIFAR-100 using the same setup as ASAM. Results for SGD, SAM, and ASAM ($^\star$) are the ones reported by \citet{kwon2021asam}.}
\begin{center}
\begin{tabular}{|l|c|}
\hline
Model & WRN-28-10\\
\hline\hline
SGD$^\star$ & $81.56_{\pm .13}$\\
\hline
SAM$^\star$ & $83.42_{\pm .04}$\\
\textbf{Lookbehind-SAM} & $\pmb{83.72_{\pm .10}}$\\
\hline
ASAM$^\star$ & $83.68_{\pm .12}$\\
\textbf{Lookbehind-ASAM} & $\pmb{84.00_{\pm .18}}$\\
\hline
\end{tabular}
\end{center}
\label{tab:generalization_asam_setup}
\end{table*}

\subsection{$m$-SAM/ASAM}
\label{sec:m_sam}

In the original SAM paper \cite{foret2021sharpnessaware}, the authors proposed $m$-SAM to reduce the mini-batch noise of the single ascent step of single-step SAM by splitting the batch size into microbatches and independently perturbing each microbatch. Then, an average of the gradients is used for the descent step. It is important to note that $m$-SAM and Lookbehind serve different purposes since Lookbehind aims to combine gradients of different ascent steps of Multistep-SAM to enhance the maximization part of SAM’s objective. However, since Lookbehind is a wrapper that can be applied to any SAM variant, we apply Lookbehind to $m$-SAM/ASAM in \cref{tab:m_sam} using ResNet-34 trained on CIFAR-10. 

\begin{table*}[h]
\caption{Generalization performance (validation acc. \%) of $m$-SAM/ASAM and Lookbehind-$m$-SAM/ASAM.}
\begin{center}
\begin{tabular}{|l|c|c|}
\hline
$m$ & $32$ & $64$\\
\hline\hline
$m$-SAM & $96.19_{\pm .09}$ & $95.99_{\pm .04}$\\
Lookbehind-$m$-SAM ($\alpha=0.2, k=2$) & $\pmb{96.45_{\pm .04}}$ & $95.91_{\pm .08}$\\
Lookbehind-$m$-SAM ($\alpha=0.5, k=2$) & $\pmb{96.61_{\pm .05}}$ & $\pmb{96.26_{\pm .15}}$\\
Lookbehind-$m$-SAM ($\alpha=0.8, k=2$) & $\pmb{96.68_{\pm .07}}$ & $\pmb{96.44_{\pm .13}}$\\
Lookbehind-$m$-SAM (adaptive $\alpha, k=2$) & $\pmb{96.66_{\pm .03}}$ & $\pmb{96.49_{\pm .05}}$\\
Lookbehind-$m$-SAM ($\alpha=0.2, k=5$) & $\pmb{96.65_{\pm .07}}$ & $\pmb{96.12_{\pm .10}}$\\
Lookbehind-$m$-SAM ($\alpha=0.5, k=5$) & $\pmb{96.53_{\pm .12}}$ & $\pmb{96.52_{\pm .04}}$\\
Lookbehind-$m$-SAM ($\alpha=0.8, k=5$) & $\pmb{96.35_{\pm .01}}$ & $\pmb{96.31_{\pm .09}}$\\
Lookbehind-$m$-SAM (adaptive $\alpha, k=5$) & $\pmb{96.35_{\pm .02}}$ & $\pmb{96.41_{\pm .05}}$\\
\hline
$m$-ASAM & $96.45_{\pm .04}$ & $96.28_{\pm .06}$\\
Lookbehind-$m$-ASAM ($\alpha=0.2, k=2$) & $\pmb{96.81_{\pm .18}}$ & $96.10_{\pm .15}$\\
Lookbehind-$m$-ASAM ($\alpha=0.5, k=2$) & $\pmb{97.00_{\pm .05}}$ & $\pmb{96.64_{\pm .11}}$\\
Lookbehind-$m$-ASAM ($\alpha=0.8, k=2$) & $\pmb{97.05_{\pm .11}}$ & $\pmb{96.54_{\pm .23}}$\\
Lookbehind-$m$-ASAM (adaptive $\alpha, k=2$) & $\pmb{96.97_{\pm .08}}$ & $\pmb{96.69_{\pm .19}}$\\
Lookbehind-$m$-ASAM ($\alpha=0.2, k=5$) & $\pmb{97.03_{\pm .02}}$ & $\pmb{96.57_{\pm .04}}$\\
Lookbehind-$m$-ASAM ($\alpha=0.5, k=5$) & $\pmb{97.01_{\pm .03}}$ & $\pmb{96.68_{\pm .07}}$\\
Lookbehind-$m$-ASAM ($\alpha=0.8, k=5$) & $\pmb{96.85_{\pm .01}}$ & $\pmb{96.76_{\pm .10}}$\\
Lookbehind-$m$-ASAM (adaptive $\alpha, k=5$) & $\pmb{96.68_{\pm .03}}$ & $\pmb{96.77_{\pm .09}}$\\\hline
\end{tabular}
\end{center}
\label{tab:m_sam}
\end{table*}

We observe that Lookbehind-$m$-SAM/ASAM consistently improves $m$-SAM/ASAM across all configurations of $\alpha$ and $k$, with the exception of $a=0.2, k=2$, which notably yields suboptimal results, as previously demonstrated in our manuscript. These findings further showcase the broad applicability of Lookbehind to additional SAM and ASAM variants.

\subsection{Speed of convergence}\label{sec:speed_convergence}

We provide an analysis of the speed of convergence between the best $\alpha$ for each $k$ for Lookbehind-SAM/ASAM with static $\alpha$ and Multistep-SAM/ASAM. The percentage of gradient computations that Lookbehind took to beat the Multistep-SAM/ASAM baselines is presented in \cref{tab:speed_convergence}, where $100\%$ refers to the number of total gradient computations performed by Multistep-SAM/ASAM.

\begin{table*}[h]
\caption{Average number of gradient computations at which the Lookbehind-SAM/ASAM variants reached the same performance as Multistep-SAM/ASAM.}
\begin{center}
\resizebox{0.9\textwidth}{!}{
\begin{tabular}{|l|c|c||c|c||c|}
\hline
Dataset & \multicolumn{2}{c||}{CIFAR-10} & \multicolumn{2}{c||}{CIFAR-100} & ImageNet\\
Model & ResNet-34 & WRN-28-2 & ResNet-50 & WRN-28-10 & ResNet-18\\
\hline\hline
Lookbehind-SAM (static $\alpha$, $k=2$) & $50\%$ & $51\%$	& $62\%$ & $50\%$ & $82\%$\\
Lookbehind-SAM (static $\alpha$, $k=5$) & $58\%$ & $53\%$	& $50\%$ & $95\%$ & -\\
Lookbehind-SAM (static $\alpha$, $k=10$) & $50\%$ & $56\%$ & $50\%$ & $24\%$ & -\\
Lookbehind-SAM (adaptive $\alpha$, $k=2$) & $50\%$ & $60\%$ & $36\%$ & $51\%$ & $96\%$\\
Lookbehind-SAM (adaptive $\alpha$, $k=5$) & $51\%$ & $63\%$ & $50\%$ & - & -\\
Lookbehind-SAM (adaptive $\alpha$, $k=10$) & $52\%$  & - & $50\%$ & - & -\\
\hline
Lookbehind-ASAM (static $\alpha$, $k=2$) & $50\%$ & $50\%$ & $50\%$ & $51\%$ & $87\%$\\
Lookbehind-ASAM (static $\alpha$, $k=5$) & $50\%$ & $50\%$ & $50\%$ & $50\%$ & -\\
Lookbehind-ASAM (static $\alpha$, $k=10$) & $51\%$ & $56\%$ & $51\%$ & $52\%$ & -\\
Lookbehind-ASAM (adaptive $\alpha$, $k=2$) & $50\%$ & $51\%$ & $50\%$ & $52\%$ & $93\%$\\
Lookbehind-ASAM (adaptive $\alpha$, $k=5$) & $50\%$ & $70\%$ & $50\%$ & $50\%$ & -\\
Lookbehind-ASAM (adaptive $\alpha$, $k=10$) & $50\%$  & -  & $52\%$ & - & -\\
\hline
\end{tabular}}
\end{center}
\label{tab:speed_convergence}
\end{table*}

We observe that both variants of Lookbehind-SAM/ASAM achieve a considerable convergence speedup compared to the Multistep-SAM/ASAM counterparts. We see a pattern where Lookbehind often matches the performance of Multistep around $50\%$ of the number of gradient computations. This is due to the learning rate scheduler used, where the learning rate is decayed by a factor of 10 in the middle of training. We note that, in the case of CIFAR-10/100, the missing values correspond to scenarios where the methods failed to converge across all 3 seeds. For ImageNet, the missing values are due to $k=5$ and $k=10$ being outside the hyperparameter range used in the paper.

\end{document}